\documentclass[lettersize,journal]{IEEEtran}
\usepackage{graphicx}
\usepackage{hyperref}
\usepackage{multirow}
\usepackage{dblfloatfix} 
\usepackage{pifont}
\usepackage[dvipsnames]{xcolor}
\usepackage{soul, url}
\usepackage{color}
\usepackage{boldline}
\usepackage{xcolor,colortbl}

\usepackage{tikz}

\usepackage{pifont}
\usepackage{caption}

\newcommand{\cmark}{\text{\ding{51}}}
\newcommand{\xmark}{\text{\ding{55}}}

\definecolor{Gray}{gray}{0.85}
\definecolor{LightCyan}{gray}{0.85}

\newcolumntype{g}{>{\columncolor{Gray}}c}

\usepackage[separate-uncertainty = true,multi-part-units = repeat]{siunitx}
\usepackage{array,amsfonts,amsmath}
\usepackage[font=small]{caption}
\usepackage{subcaption}
\ifCLASSOPTIONcompsoc
  \usepackage{cite}
\else
  \usepackage{cite}
\fi
\ifCLASSINFOpdf
\else
\fi
\begin{document}

\title{Cross-Language Speech Emotion Recognition Using Multimodal Dual Attention Transformers}


\author{Syed Aun Muhammad Zaidi, Siddique Latif, Junaid Qadir~(\IEEEmembership{Senior Member,~IEEE})
\IEEEcompsocitemizethanks{
\IEEEcompsocthanksitem  S.\ Latif is affiliated with USQ, Australia and Distributed Sensing Systems Group, Data61, CSIRO Australia. 
\IEEEcompsocthanksitem J. Qadir is with Computer Science and Engineering Department, College of Engineering, Qatar University, Doha, Qatar.
\IEEEcompsocthanksitem  
Corresponding E-mail: siddique.latif@usq.edu.au}}

\IEEEtitleabstractindextext{%
\begin{abstract}

Despite the recent progress in speech emotion recognition (SER), state-of-the-art systems are unable to achieve improved performance in cross-language settings. 
In this paper, we propose a Multimodal Dual Attention Transformer (MDAT) model to improve cross-language SER. Our model utilises pre-trained models for multimodal feature extraction and is equipped with a dual attention mechanism including graph attention and co-attention to capture complex dependencies across different modalities and achieve improved cross-language SER results using minimal target language data. In addition, our model also exploits a transformer encoder layer for high-level feature representation to improve emotion classification accuracy. In this way, MDAT performs refinement of feature representation at various stages and provides emotional salient features to the classification layer. This novel approach also ensures the preservation of modality-specific emotional information while enhancing cross-modality and cross-language interactions.  We assess our model's performance on four publicly available SER datasets and establish its superior effectiveness compared to recent approaches and baseline models.


\end{abstract}

}

\maketitle

\IEEEdisplaynontitleabstractindextext
\IEEEpeerreviewmaketitle

\section{Introduction
\label{sec:introduction}}

\IEEEPARstart{S}{peech} Emotion Recognition (SER), involves the application of machine learning (ML) systems to identify and classify emotions expressed in human speech, has recently garnered significant attention \cite{latif2020speech}. SER demonstrates a wide range of potential applications across various domains, presenting opportunities in healthcare, transportation services, forensic investigations, education, and media industries \cite{rana2019automated, zepf2020driver, tavi2020prosodic, yadegaridehkordi2019affective, vanderplaetse2020improved}. Given its complex nature, modelling human emotions in speech relies on multiple factors such as the speaker, gender, age, culture, and dialect \cite{aldeneh2021you, nediyanchath2020multi, wang2017learning, latif2018cross, laukka2014evidence}. To tackle this complexity, researchers have explored diverse ML techniques, including deep neural networks (DNNs) like Deep Belief Networks (DBN), Convolutional Neural Networks (CNNs), and Recurrent Neural Networks (RNNs) for SER \cite{hinton2006fast, lecun1989handwritten, latif2020deep, neumann2017attentive}. Among these, RNN architectures such as Long Short-Term Memory (LSTM) networks or Bidirectional LSTM (BiLSTM) combined with CNNs have been widely explored and found to be effective in modelling emotional context in speech \cite{atila2021attention, latif2019direct}. However, RNN-based models have some limitations in capturing long-term dependencies and modelling context in speech\cite{schaefer2008learning, zhang2021recurrent, liu2020subtraction}. 

\textit{Transformers} have been a major breakthrough in the field of natural language processing (NLP) and speech processing. They have revolutionized the way intricate relationships among different elements of input sequences are learned by leveraging self-attention mechanisms. Their impact has been remarkable, leading to transformative outcomes in machine translation, natural language understanding, and speech recognition tasks \cite{latif2023transformers}. This paper introduces a \textit{transformer-based multimodal technique} that uses speech and text data for cross-language SER using a dual-attention mechanism. By leveraging state-of-the-art pre-trained models for speech and text modalities, our technique enhances cross-lingual SER performance models through the use of a dual attention mechanism to capture complex multimodal representations across different languages.

Existing SER models mostly rely on within-corpus analysis, where they are trained and tested on the same dataset or language \cite{latif2021survey}. Consequently, their generalisation and applicability to different datasets or languages are hindered. This limitation is particularly evident in the case of low-resource languages like Urdu, which lack sufficient labeled data for pre-training or fine-tuning. There is a pressing need for cross-lingual models capable of performing effectively across multiple languages, leveraging the abundant data available from diverse sources. Our approach aims to fulfil this demand and contribute to the advancement of cross-lingual SER research.

A promising methodology for tackling the challenging problem of SER for low-resource languages is to employ multimodal data that capitalises on the abundant and varied cues for emotion recognition present in the complementary modalities. In this context, speech encompasses prosodic and acoustic features of emotion (e.g., pitch, intensity, and tone), while text encompasses lexical and semantic features of emotion (e.g., words, phrases, and sentiments). The inclusion of the text modality in SER is crucial as it can offer supplementary or alternative information to the speech modality in certain scenarios. For instance, when the speech signal is contaminated with noise or distortion, the text modality can provide a clearer and more dependable source of information. Additionally, the text modality can capture emotional expressions that may not be conveyed through speech, such as sarcasm, irony, or humor. The complementary nature of text and speech modalities can help enhance the accuracy of SER models, particularly for low-resource languages.

Despite the promise of combining audio and text modalities for SER, their integration into multimodal models remains largely unexplored with very few existing works \cite{poria2017review, zadeh2018multimodal}. Various text-based models, such as transformer-based ones like BERT, RoBERTa, and T5 \cite{devlin2019bert, liu2019roberta, raffel2020exploring}, can be used to learn powerful semantic representations from large-scale text corpora and transfer them to SER tasks. However, the utilization of pre-trained models for both speech and text modalities remains underexplored in SER literature, with current works \cite{poria2017review, zadeh2018multimodal} mostly relying on hand-crafted or shallow features for speech and text modalities. Our hypothesis is that by leveraging the complementary nature of pre-trained models for speech and text modalities, we can greatly enhance the generalisation of SER models for cross-language settings, paving the way for promising avenues of further investigation and improvement in this field.


This paper presents a novel multimodal model that combines RoBERTa embeddings for text and wav2vec 2.0 for speech. We utilize XLS-R \cite{babu2021xls}, a multilingual extension of wav2vec 2.0 pretrained on speech data from 128 languages, enabling cross-lingual speech representation learning. Additionally, we employ RoBERTa \cite{liu2019roberta}, a multilingual variant of BERT \cite{devlin2019bert}, for text analysis. Our model incorporates dual attention mechanism leverages (1) \textit{graph attention} to capture the complex dependencies between different parts of the modalities, such as words, phrases, or speech segments \cite{velivckovic2017graph}; as well as (2) \textit{co-attention} to capture the cross-modality interactions between speech and text modalities \cite{lu2016hierarchical}. 
By combining graph attention and co-attention, our model can effectively preserve modality-specific information while enhancing the integration of cross-modality information.




\vspace{3mm}
The \textbf{\textit{major contributions of this work}} are outlined next. 

\begin{itemize} 
\item We propose Multimodal Dual Attention Transformer (MDAT) that leverages large-scale multilingual pre-trained models for both speech and text modalities.  
These models are used for feature extraction and they can capture rich and generalised acoustic and linguistic features for different languages and domains. 
\item We propose to incorporate a dual attention mechanism to effectively model the speech and text modalities. 
By combining co-attention and graph attention, our model can preserve the modality-specific information while enhancing the cross-modality information. 
\item We propose leveraging a transformer encoder layer for high-level representation learning, which enhances cross-modality and cross-language interactions, enabling the generation of salient and generalized multimodal features to improve emotion classification.
\item  We evaluate our model on four different languages datasets to demonstrate its effectiveness and generalisation compared to existing models. 


\end{itemize}

\vspace{3mm}
\textbf{\textit{Organisation of this paper}}. Section \ref{sec:RelatedWork} provides an overview of the related work. Section \ref{sec:Methodology} describes our methodology and provides details of our proposed model architecture and the baseline model architecture. Section \ref{sec:ExperimentalSetup} provides details about the experimental setup and the datasets used. Section \ref{sec:Results} provides details of our experiments and results. Finally, Section \ref{sec:Conclusions} concludes the paper.

\section{Related Work}
\label{sec:RelatedWork}

In this paper, we present a novel framework that combines transformer-based multimodal learning with cross-language SER.
This section reviews the related work on various aspects of our technique, such as cross-lingual SER, multimodal SER, and transformer-based SER approaches. We also highlight the differences and contributions of our proposed framework compared to the existing methods.

\subsection{Cross-Language SER}

SER addresses the task of recognizing emotions from speech signals in diverse languages, with practical applications in healthcare, education, and social media \cite{rana2019automated,yadegaridehkordi2019affective,vanderplaetse2020improved}. The main challenges include limited labelled data for low-resource languages like Urdu \cite{latif2018cross}, Persian \cite{deng2020low}, or Marathi \cite{lahoti2022survey}, and domain mismatch between different speech emotion corpora \cite{godard2017very,schuller2010interspeech,schuller2013interspeech}. 
In addition, many existing SER models are trained on a single language or corpus, hence they suffer from limited applicability and generalisation. Cross-language SER presents a more realistic and challenging scenario, expanding the potential of SER for broader languages.

Recent works on cross-language SER attempt to address the challenges of data scarcity and domain mismatch by using various methods including feature selection \cite{ozseven2019novel}, domain adaptation \cite{ahn2021cross}, data augmentation \cite{kshirsagar2022cross, latif2022multitask}, and multimodal fusion \cite{shen2020wise}. Feature selection methods aim to select the most relevant and discriminative features for emotion recognition from different modalities, such as acoustic, linguistic, or prosodic features \cite{atmaja2022survey}. Domain adaptation methods aim to reduce the distribution mismatch between the source and target domains by using techniques such as adversarial learning \cite{kim2021contrastive}, transfer learning \cite{latif2018cross}, or meta-learning \cite{kuruvayil2022emotion}. Data augmentation methods aim to increase the diversity and size of the training data by applying various transformations, such as speech synthesis \cite{kshirsagar2022cross}, noise injection \cite{rana2016emotion}, or pitch shifting \cite{chatziagapi2019data}. Multimodal fusion methods aim to combine speech and text data to enhance emotion recognition accuracy by using techniques such as attention mechanisms \cite{liu2020cross}, tensor fusion \cite{zadeh2017tensor}, or graph neural networks \cite{zhou2020graph}. These methods have shown promising results on various benchmark datasets, such as IEMOCAP \cite{busso2008iemocap}, MELD \cite{poria2018meld}, or MOSI \cite{zadeh2016mosi}.


Some studies also utilise unsupervised pre-training \cite{siddhant2019unsupervised} or self-supervised pre-training \cite{medina2020self, mao2020survey} to learn general representations that can transfer across languages or domains. Additionally, meta-learning has been proposed as a promising approach to train models that can quickly adapt to new languages or tasks with few labelled examples \cite{deng2020low}. Moreover, synthetic data and data augmentation techniques, such as speech synthesis \cite{ueno2019multi,malik2023preliminary,latif2023generative} or noise injection \cite{latif2018adversarial}, can help alleviate the scarcity of labelled data for low-resource languages.

Most of the above mentioned studies also have some limitations, such as relying on hand-crafted features, requiring large amounts of labelled data, or ignoring the complex dependencies between different modalities. Therefore, there is still room for improvement in cross-language SER by developing more effective methods that can leverage the state-of-the-art pre-trained models, learn general and comprehensive representations, and capture the cross-modality interactions.

\subsection{Transformers in SER}
Transformers are a type of neural network architecture that has revolutionised NLP \cite{wolf2020transformers}. Transformers rely on self-attention mechanisms to compute representations of the input and output data without using sequence-aligned RNNs or CNNs \cite{vaswani2017attention}. Transformers can handle long-range dependencies and parallelise the computation better than RNNs or CNNs, which makes them more efficient and accurate for various NLP tasks, such as machine translation, text summarisation, question answering, and natural language understanding \cite{devlin2018bert,radford2019language}. Transformers are playing an increasingly prominent role in speech technology, particularly in the field of speech emotion recognition \cite{latif2023transformers}. They excel at harnessing the abundant semantic and acoustic information inherent in speech data to capture intricate interactions between various modalities, such as audio and text, and fully leverage the capabilities of these modalities when they are available \cite{chen2022key,abdullah2021paralinguistic}.

Several recent works use transformers for SER. For example, Chen et al. \cite{chen2022key} propose a key-sparse transformer for multimodal SER that focuses more on emotion-related information and reduces the noise from redundant information. Wagner et al. \cite{wagner2023dawn} conduct a thorough analysis of the influence of model size and pre-training data on the downstream performance of transformers for SER. They found that larger models and more diverse pre-training data lead to better results, especially for valence prediction. They also showed that transformers can learn implicit linguistic information from speech signals that can improve their valence predictions. Zenkov et al. \cite{zenkov2021transformer} build two parallel CNNs in parallel with a transformer encoder network to classify emotions from speech data. They used the RAVDESS dataset to classify emotions from one of eight classes. Li et al. \cite{li2021speech} propose a speech emotion recognition transformer (SERT) that uses a multi-head self-attention mechanism to model the temporal dynamics of speech signals. They achieved state-of-the-art results on three Engligh language benchmark datasets including IEMOCAP \cite{busso2008iemocap}, MSP-Podcast \cite{park2019msp}, and MOSI \cite{zadeh2016mosi}. Triantafyllopoulos et al. \cite{triantafyllopoulos2022probing} probed speech emotion recognition transformers for linguistic knowledge and found that they are very reactive to positive and negative sentiment content, as well as negations.


In contrast to previous works, we employ a transformer layer as part of our multimodal model for cross-language SER. A transformer encoder layer consists of two sub-layers: a multi-head self-attention layer and a feed-forward network layer \cite{vaswani2017attention}. We use the transformer layer after dual attention mechanisms to learn high-level feature extraction to improve the classification accuracy. The transformer encoder layer can enhance the representation learning by capturing the long-range dependencies and non-linear transformations within and across the modalities.


\begin{table*}[!h]
\centering
\setlength{\tabcolsep}{5pt}
\renewcommand{\arraystretch}{1.4}
\captionsetup{justification=centering}
\caption[]{
Comparison of Multimodal Emotion Recognition in Speech: Evaluating Transformer, Co-Attention, and Graph Attention Approaches for Same Corpus and Cross-Language Analysis.
}
\begin{tabular}{|c|c|c|c|c|c|}
\hline
\multirow{2}{*}{\textbf{Work (Year)}} & \multicolumn{2}{c|}{\textbf{Evaluations}} & \multirow{2}{*}{\textbf{\begin{tabular}[c]{@{}c@{}}
Transformer Based\end{tabular}}} & \multirow{2}{*}{\textbf{\begin{tabular}[c]{@{}c@{}}
Co-Attention\end{tabular}}} & \multicolumn{1}{c|}{\multirow{2}{*}{\textbf{\begin{tabular}[c]{@{}c@{}}
Graph Attention\end{tabular}}}} \\ \cline{2-3}
& \textbf{\begin{tabular}[c]{@{}c@{}}
Same Corpus\end{tabular}} & \textbf{\begin{tabular}[c]{@{}c@{}}
Cross Language\end{tabular}} & & & \multicolumn{1}{c|}{} \\ \hline
Chen et al. (2019) \cite{chen2019complementary} 
& $\cmark{}$ 
& $\xmark{}$ 
& $\xmark{}$ 
& $\xmark{}$ 
& $\xmark{}$ 
\\ \hline
Huang et al. (2020) \cite{huang2020multimodal} 
& $\cmark{}$ 
& $\xmark{}$ 
& $\cmark{}$ 
& $\xmark{}$ 
& $\xmark{}$ 
\\ \hline
Siriwardhana et al. (2020) \cite{siriwardhana2020multimodal} 
& $\cmark{}$ 
& $\xmark{}$ 
& $\cmark{}$ 
& $\xmark{}$ 
& $\xmark{}$ 
\\ \hline
Su et al. (2020) \cite{su2020improving} 
& $\cmark{}$ 
& $\xmark{}$ 
& $\xmark{}$ 
& $\xmark{}$ 
& $\cmark{}$ 
\\ \hline
Zhou et al. (2020) \cite{zhou2020graph} 
& $\cmark{}$ 
& $\xmark{}$ 
& $\cmark{}$ 
& $\xmark{}$ 
& $\cmark{}$ 
\\ \hline
Wang et al. (2021) \cite{wang2021learning} 
& $\cmark{}$ 
& $\xmark{}$ 
& $\cmark{}$ 
& $\xmark{}$ 
& $\xmark{}$ 
\\ \hline
Zheng et al. (2021) \cite{zheng2021fused}  &$\cmark{}$
&$\cmark{}$
&$\cmark{}$
&$\xmark{}$
&$\xmark{}$
\\ \hline
Chen et al. (2022) \cite{chen2022key}  &$\cmark{}$
&$\xmark{}$
&$\cmark{}$
&$\cmark{}$
&$\xmark{}$
\\ \hline
Kim et al. (2022) \cite{kim2022representation}  &$\cmark{}$ 
&$\xmark{}$
&$\xmark{}$
&$\xmark{}$
&$\cmark{}$
\\ \hline
Zhang et al. (2022) \cite{zhang2022transformer}  &$\cmark{}$ 
&$\xmark{}$
&$\cmark{}$
&$\cmark{}$
&$\xmark{}$
\\ \hline
Guo et al. (2022) \cite{guo2022emotion}  &$\cmark{}$ 
&$\xmark{}$
&$\cmark{}$
&$\cmark{}$
&$\xmark{}$
\\ \hline
Yang et al. (2022) \cite{yang2022self}  &$\cmark{}$ 
&$\xmark{}$
&$\xmark{}$
&$\xmark{}$
&$\xmark{}$
\\ \hline
Wang et al. (2023) \cite{wang2023multimodal}  &$\cmark{}$ 
&$\xmark{}$
&$\cmark{}$
&$\xmark{}$
&$\xmark{}$
\\ \hline
Dutta et al. (2023) \cite{dutta2023hcam}  &$\cmark{}$ 
&$\xmark{}$
&$\xmark{}$
&$\xmark{}$
&$\xmark{}$
\\ \hline
\textit{This work} (2023)  &$\cmark{}$
&$\cmark{}$
&$\cmark{}$
&$\cmark{}$
&$\cmark{}$
\\ \hline
\end{tabular}
\label{table:research_gap}
\end{table*}

\subsection{Multimodal SER}

Multimodal SER involves the identification of human affective states from speech data using multiple sources of information, including audio, text, visual, and paralinguistic features. This approach has demonstrated superior accuracy and robustness compared to single-modality approaches \cite{xie2021robust, schlegel2021training, wang2022multi, tsai2019multimodal}. Nevertheless, there are several challenges associated with multimodal emotion recognition, such as feature extraction difficulties \cite{liu2021multi}, feature alignment complexities \cite{liu2022multi}, fusion techniques \cite{ho2020multi}, dealing with missing or noisy data \cite{liu2021multi}. Therefore, we need more advanced methods that can exploit the information from multimodal data in an effective way. This can enhance multimodal SER by integrating multiple modalities, and provide us with a richer understanding of human emotions.

Several methods are proposed for multimodal SER to improve performance by using pre-trained models for feature extraction. Makiuchi et al. \cite{morais2022speech} propose a cross-representation speech model that combines self-supervised high-level features extracted from raw audio waveforms with text-based features extracted with Transformer-based models. They achieve state-of-the-art results on the IEMOCAP dataset using a score fusion approach, but their method may require fine-tuning for different languages or domains. Tang et al. \cite{tang2022multimodal} propose a feature fusion method based on facial expression and speech using attention mechanisms, which showed improved accuracy on the RAVDESS dataset, but may not generalise well to other datasets or modalities. Yoon et al. \cite{yoon2018multimodal} propose a deep dual recurrent encoder model that uses text data and audio signals simultaneously to better understand speech data. Their model outperforms previous state-of-the-art methods in emotion classification on the IEMOCAP dataset but may not capture long-term dependencies or context information and depend on the quality of the text transcripts.

Different recent works have proposed a novel fusion techniques for multimodal SER that use hybrid transformer models \cite{chen2022key,wagner2023dawn,jin2021hybrid}. Hybrid transformer models are models that combine different types of transformer architectures, such as encoder-decoder, encoder-only, or decoder-only, to achieve better performance on multimodal tasks \cite{chen2022key}. For example, Chen et al. \cite{chen2022key} propose a key-sparse attention model that uses an encoder-decoder transformer to fuse multimodal data in a more efficient and adaptive manner. They used a key-sparse attention mechanism to reduce the computational complexity and memory consumption of the self-attention mechanism by selecting a subset of keys for each query. They show that their model can achieve state-of-the-art results on various multimodal tasks, including multimodal SER \cite{chen2022key}. Wagner et al. \cite{wagner2023dawn} propose a progressive fusion model that uses an encoder-only transformer to fuse multimodal data in a more refined and iterative manner. They use a progressive fusion technique that introduces backward connections between different stages of fusion to preserve modality-specific information while enhancing cross-modality interactions. Jin et al. \cite{jin2021hybrid} propose a hybrid transformer model that uses a decoder-only transformer to fuse multimodal data in a more flexible and scalable manner. They use a hybrid transformer architecture that combines the advantages of autoregressive and non-autoregressive transformers to generate multimodal outputs with high quality and efficiency.  They evaluate their model on various multimodal tasks, including multimodal SER \cite{jin2021hybrid}.

\subsection{Comparing Proposed Model with Related Works}
We will now compare our proposed MDAT model with previous methods. Table \ref{table:research_gap} summarises the existing works on multimodal techniques and their evaluations. Upon reviewing the previous studies, it becomes evident that most previous multimodal methods solely evaluated their models on the same corpus setting.
Only a few of them utilised transformer-based multimodal models for cross-language SER. One of the major challenges in cross-language SER is the scarcely of labelled datasets in different languages. Here we address this issue by proposing a transformer-based multimodal model that utilises dual attention to learn generalised multimodal representations and can improve the performance of few shots target samples adaptation. 

To summarise, the key aspects that distinguish our proposed model are as follows:

\begin{itemize} 
\item In contrast to previous works, MDAT uses multilingual pre-trained models to get feature embeddings on each of our modalities.
Using these models can improve the performance and generalizability of our model compared to the previous methods \cite{chen2019complementary,huang2020multimodal,zhou2020graph}.
\item MDAT leverages graph attention and co-attention networks to capture the fine-grained dependencies between modalities and preserve the modality-specific information in contrast to the single-type attention layer used in previous studies \cite{chen2019complementary,huang2020multimodal,dutta2023hcam}.


\item MDAT uses a transformer encoder layer to learn high-level feature representation to improve the multimodal cross-language SER. 
The transformer encoder layer help improves the performance of our model compared to the previous methods that use recurrent or graph-based layers, such as BiLSTM \cite{chen2019complementary}, BiGRU \cite{zhou2020graph}, or HCAM \cite{dutta2023hcam}. 
\end{itemize}

\section{Proposed Model Architecture}
\label{sec:Methodology}

\begin{figure}[!h]
\centering
\includegraphics[width=.8\linewidth]{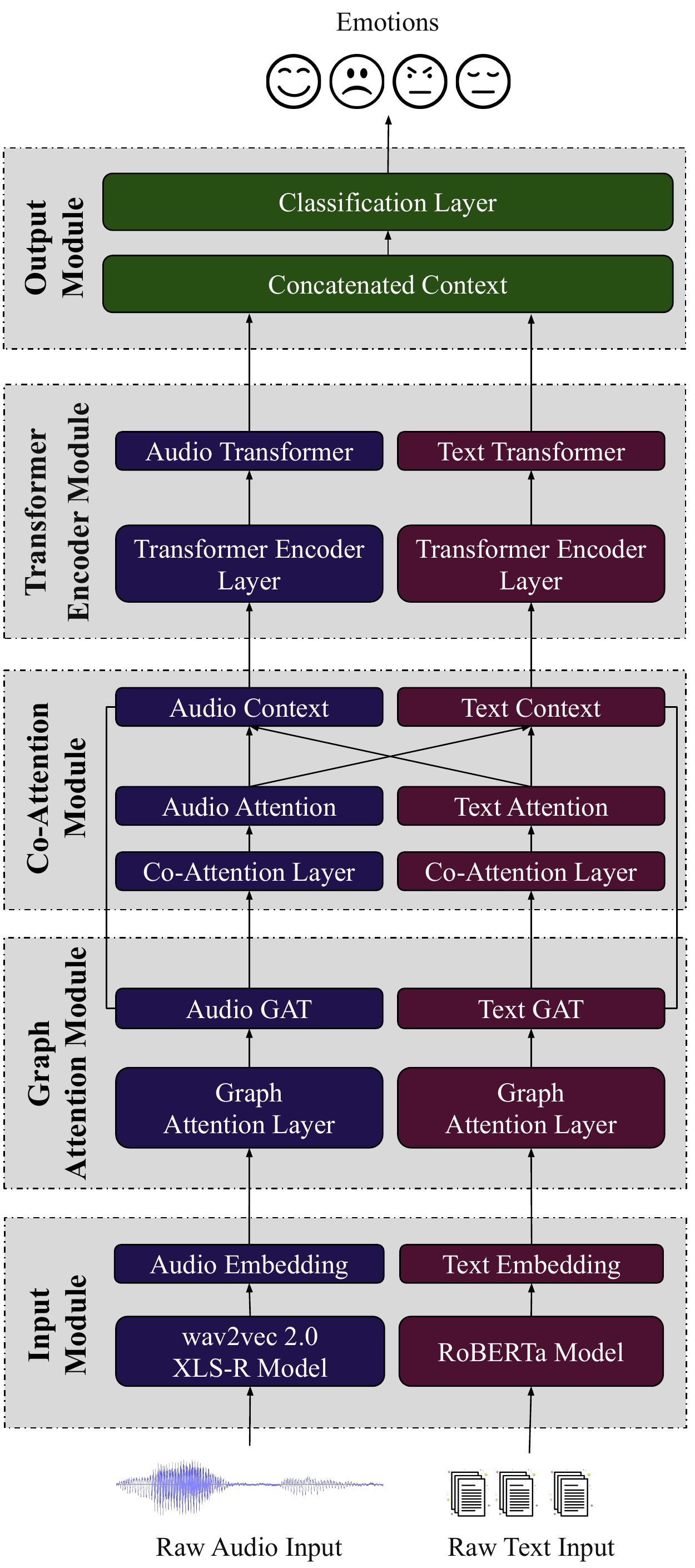}
\caption{In this work, we propose a Multimodal Dual Attention Transformer (MDAT), which uses graph attention networks to compute weighted sums of the feature vectors according to their similarity and importance, and co-attention networks to capture cross-modal interactions and dependencies between the audio and text modalities. The co-attention networks apply dense layers, activations, and element-wise multiplications to the graph attention vectors to obtain co-attention vectors. The co-attention vectors are then processed by a transformer encoder, which consists of self-attention and feed-forward sub-layers. The classification  layer concatenates the audio and text transformer encoder vectors and applies a series of dense and dropout layers to produce the output probabilities.
}
\label{fig:model_architecture}
\end{figure}

The overall architecture of our model is shown in Figure~\ref{fig:model_architecture}. Our proposed Multimodal Dual Attention Transformer (MDAT) leverages pre-trained models, RoBERTa and wav2vec 2.0 based XLS-R, for feature embedding creation from each modality. The model architecture exploits the dual attention mechanisms to capture the complex dependencies between different parts of each modality. We also use a BiLSTM-based multimodal model as a baseline to compare with our proposed model and show the effectiveness of our proposed framework. This section describes the details of our proposed model architecture for multimodal SER. 

\subsection{Input Feature Layers}
The input layer takes speech and text features from pre-trained models XLS-R and RoBERTa, respectively. Let $T_s$ and $T_t$ be the sequence lengths and $D_s$ and $D_t$ be the feature dimensions for speech and text modalities. The speech input layer has a shape of $(T_s, D_s)$, while the text input layer has a shape of $(T_t, D_t)$. Since $D_s \neq D_t$, we use a convolutional layer with kernel size 1 to transform the text features to have the same dimensionality as the speech features, i.e., $D_s$. The convolutional layer is defined as:
\begin{equation}
\mathbf{x}_t = \text{Conv1D}(\mathbf{e}_t; W_c, b_c)
\end{equation}
where $\mathbf{e}_t \in \mathbb{R}^{T_t \xmark{} D_t}$ is the input text feature matrix, $W_c \in \mathbb{R}^{D_s \times D_t}$ and $b_c \in \mathbb{R}^{D_s}$ are the convolutional weights and bias, and $\mathbf{x}_t \in \mathbb{R}^{T_t \times D_s}$ is the output text feature matrix.

To ensure that the speech and text features have the same length, we pad the shorter sequence and crop the longer ones. This is necessary for applying the subsequent dual attention and transformer layers that require aligned inputs. We denote the padded or cropped speech and text features as $\tilde{\mathbf{x}}_s$ and $\tilde{\mathbf{x}}_t$, respectively. We pass these features to the graph attention layers to learn the dependencies between audio and text. We describe the function of the graph attention layer next. 


\subsection{Graph Attention Layer}
The graph attention layer is used to capture the complex dependencies between different parts of the audio and text modalities and generate updated features for each modality. The motivation behind using graph attention is to model the input data as a graph where the nodes represent the features and the edges represent the similarity between features. By applying a graph attention operation on the input features, we can aggregate the information from the neighbouring nodes using attention weights and update the features accordingly. This technique is useful for capturing complex dependencies between different parts of the audio and text modalities that may not be captured by other methods.

We implement the graph attention layer based on the original paper by Velivckovic et al. (2017) \cite{velivckovic2017graph}. The graph attention layer consists of two steps:
\begin{enumerate}
    \item Computing the attention weights between each pair of features in the same modality.
    \item Multiplying the attention weights with the original features to generate updated features.
\end{enumerate}
We apply a graph attention layer separately for each modality. The graph attention layer is defined as:
\begin{align}
\mathbf{Z} &= [\tilde{\mathbf{x}}_s; \tilde{\mathbf{x}}_t] \\
\mathbf{H} &= \mathbf{Z}W \\
\mathbf{A} &= \text{LeakyReLU}(\mathbf{H}\mathbf{a}) \\
\mathbf{\alpha} &= \text{softmax}(\mathbf{A}) \\
\hat{\mathbf{x}}_g &= \mathbf{\alpha}\mathbf{H}
\end{align}
where $\mathbf{Z} \in \mathbb{R}^{(T_s + T_t) \times D_s}$ is the input feature matrix for both modalities, $W \in \mathbb{R}^{D_s \times U}$ is the weight matrix, $\mathbf{a} \in \mathbb{R}^{U}$ is the attention vector, $\alpha \in \mathbb{R}^{(T_s + T_t) \times (T_s + T_t)}$ is the graph attention matrix, and $\hat{\mathbf{x}}_g
\in \mathbb{R}^{(T_s + T_t) \times U}$ is the updated feature matrix for both modalities. The graph attention layer updates the features for each modality by aggregating the information from both modalities. This allows our model to capture the complex dependencies between different parts of the audio and text modalities and generate a more emotionally salient multimodal representation.

\subsection{Co-Attention Layer}
The co-attention layer is used to align the audio and text modalities and generate attended features for each modality. The motivation behind using co-attention is to compute the semantic similarity between each pair of features from both modalities and produce an attention matrix. This attention matrix is then used to generate context vectors for each modality by multiplying it with the original features. This technique is useful for capturing cross-modality interactions and generating a comprehensive representation of the input data.

We implement the co-attention layer based on the original paper by Lu et al. (2016) \cite{lu2016hierarchical}, but with some modifications to adapt it to our task and data. The co-attention layer consists of two steps:
\begin{enumerate}
    \item the attention weights between each pair of features from both modalities.
    \item Multiplying the attention weights with the original features to generate context vectors.
\end{enumerate}
We apply a co-attention layer to align the audio and text modalities. The co-attention layer is defined as:
\begin{align}
\mathbf{H}_s &= \text{Dense}(\tilde{\mathbf{x}}_s; W_s, b_s) \\
\mathbf{H}_t &= \text{Dense}(\tilde{\mathbf{x}}_t; W_t, b_t) \\
\mathbf{C} &= \mathbf{H}_s^T\mathbf{H}_t \\
\mathbf{\alpha}_s &= \text{softmax}(\mathbf{C}) \\
\mathbf{\alpha}_t &= \text{softmax}(\mathbf{C}^T) \\
\mathbf{x}_s' &= \tilde{\mathbf{x}}_s\mathbf{\alpha}_s \\
\mathbf{x}_t' &= \tilde{\mathbf{x}}_t\mathbf{\alpha}_t
\end{align}
where $\mathbf{H}_s \in \mathbb{R}^{T_s \times D_s}$ and $\mathbf{H}_t \in \mathbb{R}^{T_t \times D_s}$ are the transformed features for each modality using dense layers with weights $W_s, W_t$ and bias $b_s, b_t$, $\mathbf{C} \in \mathbb{R}^{T_s \times T_t}$ is the co-attention matrix between audio and text modalities, $\mathbf{\alpha}_s \in \mathbb{R}^{T_s \times T_t}$ and $\mathbf{\alpha}_t \in \mathbb{R}^{T_t \times T_s}$ are the attention matrices for each modality, $\mathbf{x}_s' \in \mathbb{R}^{T_s \times D_s}$ and $\mathbf{x}_t' \in \mathbb{R}^{T_t \times D_s}$ are the attended features for each modality. 
To further enhance the representation, we concatenate the input features to the co-attention layer and attended features to form a richer representation for each modality. This enables our model to capture both the modality-specific features and the cross-modality interactions in a comprehensive manner. The concatenated features are defined as:
\begin{align}
\hat{\mathbf{x}}_s &= [\tilde{\mathbf{x}}_s; \mathbf{x}_s'] \\
\hat{\mathbf{x}}_t &= [\tilde{\mathbf{x}}_t; \mathbf{x}_t']
\end{align}
where $\hat{\mathbf{x}}_s
\in \mathbb{R}^{T_s \times 2D_s}$ and $\hat{\mathbf{x}}_t
\in \mathbb{R}^{T_t \times 2D_s}$ are the concatenated features for each modality. The graph attention features for both audio and text modalities are directly connected to the co-attention layer, where they are transformed by dense layers and used to compute the co-attention weights. The co-attention weights are then multiplied element-wise with the graph attention features to generate the co-attended features for each modality.

The main differences between our implementation and the original paper \cite{lu2016hierarchical} are that we use a dense layer to transform the features before computing the co-attention matrix, and we use a softmax function instead of a sigmoid function to normalise the attention weights. These modifications help improve the performance of our model and make it more suitable for multimodal cross-language SER.

\subsection{Transformer Encoder Layer and Classification Layer}
The transformer encoder layer is used to process both audio and text contexts and generate transformer outputs for each modality. The motivation behind using a transformer encoder is to use self-attention and feed-forward networks to capture the long-range dependencies and semantic information within each modality. This technique is useful for enhancing the representation of each modality and making it more suitable for classification. The transformer encoder layer consists of the following two sub-layers. First is the multi-head self-attention sub-layer, which computes the attention weights between each pair of features in the same modality and generates context vectors. Second is the feed-forward sub-layer, which applies a linear transformation and an activation function to the context vectors.  We apply a transformer encoder layer separately for each modality.  

The output classification layer concatenates the audio and text transformer outputs and applies a dense layer with softmax activation to predict the emotion class of the input. This layer computes the final output probabilities based on both modalities. 



\subsection{Baseline Model Architecture}
\label{methodology:baseline-architechture}
The baseline model architecture is a simple bidirectional LSTM (BiLSTM) model that concatenates the audio and text features after applying BiLSTM layers to each modality. We choose this model as a baseline because it is widely used for sequence modeling tasks and it does not use any sophisticated fusion technique. The input layer takes speech and text features as inputs. The speech and text features have different lengths and dimensions, so we need to align them before concatenating them. The BiLSTM layer is applied to each modality separately to capture the sequential information. The BiLSTM layer consists of two LSTM sub-layers that process the input sequence from both forward and backward directions and concatenate their outputs. This allows the model to capture both the past and future context of each feature. The concatenation layer concatenates the audio and text features along the feature dimension to combine both modalities into a single feature vector. This feature vector contains both the modality-specific and the cross-modality information. The final layers consist of a dense layer with regularization and dropout, and an output layer with softmax activation to predict the emotion class of the input. 

\section{Experimental Setup}
\label{sec:ExperimentalSetup}

\subsection{Datasets}
\label{sec:data}
In order to assess the effectiveness of our multimodal framework, we have employed four commonly used datasets including IEMOCAP, EMODB, EMOVO, and URDU to cover a variety of languages for evaluations. The details of these corpora is presented below. 

\subsubsection{IEMOCAP} IEMOCAP \cite{busso2008iemocap} is a widely used and publicly available multimodal dataset of emotional speech and text in English. This corpus was collected by researchers from the University of Southern California and consists of 12 hours of audiovisual data from 10 actors (5 male and 5 female) in scripted and improvised scenarios. The actors performed a range of emotions such as anger, happiness, sadness, frustration, and neutrality. Each utterance has been labelled by multiple human raters using both categorical and dimensional labels. Specifically, we employ a subset of the dataset, both the audio and text transcripts of 800 utterances, encompassing four emotions: angry, happy, neutral, and sad, although the exact number of utterances used may vary across different experiments. 


\subsubsection{EMODB}
EMODB \cite{burkhardt2005database} is a widely recognised and publicly available emotional dataset in German. The dataset was recorded by researchers at the Institute of Communication Science, Technical University Berlin, and consists of audio recordings of ten professional speakers expressing seven different emotions in 10 German sentences. For our study, we have selected a total of 420 utterances, including 127 angry, 143 sad, 79 neutral, and 71 happy expressions, for categorical cross-language SER analysis.


\subsubsection{EMOVO}
EMOVO \cite{costantini2014emovo} is an Italian emotional speech corpus that was developed by researchers at the Laboratory of Language Technologies at Roma Tre University. The dataset consists of 14 sentences delivered by six actors, with an equal representation of three male and three female voices. These sentences encompass seven distinct emotional states: anger, disgust, fear, joy, sadness, surprise, and neutral. These emotions represent the widely recognised ``Big Six'' emotions that are commonly used in emotional speech research. For our study, we have selected 336 utterances, with 84 utterances each for angry, happy, neutral, and sad emotions, for categorical cross-language SER analysis.

\subsubsection{URDU}
URDU \cite{latif2018cross} is an emotional speech dataset in Urdu which contains 400 utterances representing four fundamental emotions: angry, happy, neutral, and sad. There are 38 speakers (27 male and 11 female) who were selected randomly from Urdu talk shows on YouTube. In this work, we use both audio and text transcripts of the utterances for multimodal SER as done in \cite{latif2022self}. We use EmulationAI API\footnote{https://emulationai.com/call-insights/} for textual transcript generation from the given Urdu audio. We use all 400 utterances for angry (100), happy (100), neutral (100), and sad (100) emotions.

\section{Experiments and Results}
\label{sec:Results}
In this section, we present the results of our proposed MDAT model on various SER tasks. We compare our model with baselines and different state-of-the-art methods on different. We conduct two types of experiments: (1) \textit{within corpus experiments}, where we evaluate our model on the same corpus as the training data, using standard train-test splits; and (2) \textit{cross language experiments}, where we evaluate our model on a different language than the training data. We report the unweighted accuracy (UA) as the performance metric for all the models. 


\begin{table}[]
\centering
\captionsetup{justification=centering}
\caption{Comparison of results (UA) Results from within-corpus setting.
}
\label{tab:same_corpus}
\setlength{\tabcolsep}{5pt}
\renewcommand{\arraystretch}{1.2}
\begin{tabular}{|l|cccc|}
\hline
\multirow{2}{*}{\textbf{Datasets}} 
& \multicolumn{4}{c|}{\textbf{ Models performance UA (\%)}}  \\ \cline{2-5} 
& \multicolumn{1}{c|}{\textbf{\begin{tabular}[c]{@{}c@{}} Baseline\end{tabular}}} 
& \multicolumn{1}{c|}{\textbf{\begin{tabular}[c]{@{}c@{}}SAFRLM \cite{yang2022self}\end{tabular}}} 
& \multicolumn{1}{c|}{\textbf{\begin{tabular}[c]{@{}c@{}}HCAM \cite{dutta2023hcam} \end{tabular}}} 
& \textbf{\begin{tabular}[c]{@{}c@{}}MDAT\end{tabular}} \\ \hline
\begin{tabular}[c]{@{}c@{}}IEMOCAP\\ (4 Classes)\end{tabular} 
& \multicolumn{1}{c|}{63.33 \iffalse $\pm$ \fi}   
& \multicolumn{1}{c|}{72.14 \iffalse $\pm$ \fi} 
& \multicolumn{1}{c|}{73.67 \iffalse $\pm$ \fi} 
& \textbf{75.58} \\ \hline
\begin{tabular}[c]{@{}c@{}}EMODB\\ (7 Classes)\end{tabular}   
& \multicolumn{1}{c|}{81.00 \iffalse $\pm$ \fi}   
& \multicolumn{1}{c|}{73.43 \iffalse $\pm$ \fi} 
& \multicolumn{1}{c|}{83.16 \iffalse $\pm$ \fi} 
& \textbf{84.50} \\ \hline
\begin{tabular}[c]{@{}c@{}}URDU\\ (4 Classes)\end{tabular}    
& \multicolumn{1}{c|}{91.13 \iffalse $\pm$ \fi}   
& \multicolumn{1}{c|}{84.73 \iffalse $\pm$ \fi} 
& \multicolumn{1}{c|}{91.56 \iffalse $\pm$ \fi} 
& \textbf{94.33} \\ \hline
\begin{tabular}[c]{@{}c@{}}EMOVO\\ (6 Classes)\end{tabular}   
& \multicolumn{1}{c|}{72.25 \iffalse $\pm$ \fi}   
& \multicolumn{1}{c|}{67.50 \iffalse $\pm$ \fi} 
& \multicolumn{1}{c|}{76.66 \iffalse $\pm$ \fi} 
& \textbf{82.81} \\ \hline
\end{tabular}
\end{table}

\subsection{Within Corpus Experiments} 
\label{within}

\begin{table}[]
\scriptsize
\captionsetup{justification=centering}
\caption{Comparison of results (UA) Results from within-corpus setting on IEMOCAP Dataset.}
\label{tab:same_corpus_models}
\centering
\setlength{\tabcolsep}{5pt}
\renewcommand{\arraystretch}{1.4}
\begin{tabular}{|c|cc|c|}
\hline
\multirow{2}{*}{\textbf{Models}} & \multicolumn{2}{c|}{\textbf{Feature Extraction}}         & \multirow{2}{*}{\textbf{\begin{tabular}[c]{@{}c@{}} UA (\%)\end{tabular}}} \\ \cline{2-3}
                                 & \multicolumn{1}{c|}{\textbf{Audio}}      & \textbf{Text} &                                                                                                    \\ \hline
Baseline                         & \multicolumn{1}{c|}{XLS-R}         & RoBERTa       & 63.33                                                                                              \\ \hline
Yoon et al. (2018) \cite{yoon2018multimodal}              & \multicolumn{1}{c|}{openSMILE}           & Google ASR    & 48.70                                                                                              \\ \hline
Xu et al. (2019) \cite{xu2019learning}                & \multicolumn{1}{c|}{MFCC}                & BiLSTM        & 69.50                                                                                              \\ \hline
Sebastian et al. (2019) \cite{sebastian2019fusion}          & \multicolumn{1}{c|}{openSMILE}           & CNN           & 59.30                                                                                              \\ \hline
Chen et al. (2020) \cite{chen2020multi}           & \multicolumn{1}{c|}{log-mel spectrograms}                 & ALBERT         & 72.82                                                                                              \\ \hline
Krishna et al. (2020) \cite{krishna2020multimodal}           & \multicolumn{1}{c|}{CNN}                 & GloVe         & 72.82                                                                                              \\ \hline
Sun et al. (2021) \cite{sun2021multimodal}               & \multicolumn{1}{c|}{CNN + LSTM}          & BiLSTM        & 56.00                                                                                              \\ \hline
Lian et al. (2021)  \cite{lian2021ctnet}              & \multicolumn{1}{c|}{openSMILE}           & Lexical Features             & 67.60                                                                                              \\ \hline
Kumar et al. (2021) \cite{kumar2021towards}             & \multicolumn{1}{c|}{ \begin{tabular}[c]{@{}l@{}} MFCC + chroma +\\ mel\end{tabular}} & BERT          & 75.00                                                                                              \\ \hline
Yang et al. (2022) \cite{yang2022self}              & \multicolumn{1}{c|}{COVAREP}             & GloVe         & 72.14                                                                                              \\ \hline
Wang et al. (2023) \cite{wang2023multimodal}              & \multicolumn{1}{c|}{Librosa}             & TFIDF         & 75.08                                                                                              \\ \hline
Dutta et al. (2023) \cite{dutta2023hcam}             & \multicolumn{1}{c|}{wav2vec 2.0}         & RoBERTa       & 73.67                                                                                              \\ \hline
MDAT                         & \multicolumn{1}{c|}{XLS-R}         & RoBERTa       & \textbf{75.58}                                                                                              \\ \hline
\end{tabular}
\end{table}

This section evaluates the results of our proposed model on the same corpus as the training data, using standard train-test splits. We use four datasets with different languages and emotion classes: IEMOCAP (English), URDU (Urdu), EMOVO (Italian), and EMODB (German). We compare our model with three baseline models: a simple bidirectional LSTM-based baseline model that we implemented using a simple fusion of audio and text features (see \ref{methodology:baseline-architechture}), SAFRLM \cite{yang2022self} that uses a self-adjusting fusion representation learning model for unaligned text-audio sequences, and HCAM \cite{dutta2023hcam} that uses a hierarchical cross attention model for multimodal emotion recognition. We report the accuracy as the performance metric for all the models. Table \ref{tab:same_corpus} shows the results of the same corpus experiments.

Our proposed model outperformed the baseline models, SAFRLM, and HCAM models on the IEMOCAP dataset, effectively capturing multimodal information and emotion dynamics. In contrast to other approaches, our proposed model also achieve improved classification accuracy on all the datasets, which demonstrating its generalisation to different languages. The SAFRLM model also performs well, handling unaligned text-audio sequences. The HCAM model shows slight improvement over the baseline model, leveraging cross-attention for inter-modal interactions. However, both HCAM and the baseline model fall short compared to our proposed model and SAFRLM, indicating limited exploitation of multimodal features and temporal dependencies. The baseline model performs the poorest, lacking cross-modality interaction consideration in its simple fusion of audio and text features.

Results reported in Table \ref{tab:same_corpus_models} demonstrate the effectiveness and robustness of our proposed model for speech emotion recognition in different languages and emotion classes. Our model consistently achieves the highest accuracy on all the datasets, surpassing the baseline and the state-of-the-art models by a large margin. Our model leverages the power of wav2vec 2.0's XLS-R and RoBERTa for feature extraction, as well as the dual attention module, that comprehends the intricate pattern from each modality using graph attention and fuses the cross-modal information from the subsequent co-attention layer.

\begin{table*}[]
\centering
\captionsetup{justification=centering}
\caption{Results of cross-language experiments evaluating the performance of various multimodal models, using unweighted accuracy as the assessment metric.}
\setlength{\tabcolsep}{5pt}
\renewcommand{\arraystretch}{1.4}
\begin{tabular}{|l|l|cccc|}
\hline
\multirow{2}{*}{\textbf{Source}}    & \multirow{2}{*}{\textbf{Target}} & \multicolumn{4}{c|}{\textbf{Models performance UA (\%)}}  \\ \cline{3-6} 
 &    & \multicolumn{1}{c|}{\textbf{\begin{tabular}[c]{@{}c@{}} Baseline\\ multimodal Model\end{tabular}}} 
 & \multicolumn{1}{c|}{\textbf{\begin{tabular}[c]{@{}c@{}}SAFRLM \cite{yang2022self}\end{tabular}}} 
 & \multicolumn{1}{c|}{\textbf{\begin{tabular}[c]{@{}c@{}}HCAM \cite{dutta2023hcam} \end{tabular}}} 
 & \textbf{\begin{tabular}[c]{@{}c@{}}MDAT\end{tabular}} \\ \hline
\multirow{3}{*}{\begin{tabular}[c]{@{}c@{}}IEMOCAP (English)\end{tabular}} 
& EMODB (German) 
& \multicolumn{1}{c|}{40.78 \iffalse $\pm$ \fi} 
& \multicolumn{1}{c|}{39.61 \iffalse $\pm$ \fi}  
& \multicolumn{1}{c|}{38.63 \iffalse $\pm$ \fi} 
& \textbf{42.48} \\ \cline{2-6} 
 & EMOVO (Italian)    
 & \multicolumn{1}{c|}{73.46 \iffalse $\pm$ \fi } 
 & \multicolumn{1}{c|}{69.60 \iffalse $\pm$ \fi }  
 & \multicolumn{1}{c|}{73.52 \iffalse $\pm$ \fi } 
 & \textbf{85.51} \\ \cline{2-6} 
 & URDU (Urdu) 
 & \multicolumn{1}{c|}{63.18 \iffalse $\pm$ \fi } 
 & \multicolumn{1}{c|}{50.07 \iffalse $\pm$ \fi } 
 & \multicolumn{1}{c|}{63.17 \iffalse $\pm$ \fi } 
 & \textbf{64.43} \\ \hline
\multirow{3}{*}{\begin{tabular}[c]{@{}c@{}}EMODB (German)\end{tabular}}    
& IEMPCAP (English)  
& \multicolumn{1}{c|}{49.23 \iffalse $\pm$ \fi } 
& \multicolumn{1}{c|}{50.77} 
& \multicolumn{1}{c|}{55.03 \iffalse $\pm$ \fi }    
& \textbf{55.55} \\ \cline{2-6} 
 & EMOVO (Italian)    
 & \multicolumn{1}{c|}{50.54} 
 & \multicolumn{1}{c|}{52.25 \iffalse $\pm$ \fi } 
 & \multicolumn{1}{c|}{51.13 \iffalse $\pm$ \fi }    
 & \textbf{56.25} \\ \cline{2-6} 
 & URDU (Urdu) 
 & \multicolumn{1}{c|}{72.46 \iffalse $\pm$ \fi } 
 & \multicolumn{1}{c|}{70.00 \iffalse $\pm$ \fi } 
 & \multicolumn{1}{c|}{72.98 \iffalse $\pm$ \fi }    
 & \textbf{73.23} \\ \hline
\multirow{3}{*}{\begin{tabular}[c]{@{}c@{}}URDU (Urdu)\end{tabular}}  
& IEMPCAP (English)  
& \multicolumn{1}{c|}{47.96 \iffalse $\pm$ \fi } 
& \multicolumn{1}{c|}{44.75 \iffalse $\pm$ \fi } 
& \multicolumn{1}{c|}{50.55 \iffalse $\pm$ \fi }    
& \textbf{58.32} \\ \cline{2-6} 
 & EMODB (German) 
 & \multicolumn{1}{c|}{65.35 \iffalse $\pm$ \fi } 
 & \multicolumn{1}{c|}{63.40 \iffalse $\pm$ \fi } 
 & \multicolumn{1}{c|}{72.71 \iffalse $\pm$ \fi }    
 & \textbf{75.31} \\ \cline{2-6} 
 & EMOVO (Italian)    
 & \multicolumn{1}{c|}{63.55 \iffalse $\pm$ \fi } 
 & \multicolumn{1}{c|}{65.36 \iffalse $\pm$ \fi } 
 & \multicolumn{1}{c|}{63.55 \iffalse $\pm$ \fi }    
 & \textbf{67.66} \\ \hline
\multirow{3}{*}{\begin{tabular}[c]{@{}c@{}}EMOVO (Italian)\end{tabular}}   
& IEMPCAP (English)  
& \multicolumn{1}{c|}{57.25 \iffalse $\pm$ \fi } 
& \multicolumn{1}{c|}{55.17 \iffalse $\pm$ \fi } 
& \multicolumn{1}{c|}{49.09 \iffalse $\pm$ \fi }    
& \textbf{59.96} \\ \cline{2-6} 
 & EMODB (German) 
 & \multicolumn{1}{c|}{64.60 \iffalse $\pm$ \fi } 
 & \multicolumn{1}{c|}{64.28 \iffalse $\pm$ \fi } 
 & \multicolumn{1}{c|}{75.22 \iffalse $\pm$ \fi }    
 & \textbf{81.60} \\ \cline{2-6}   
 & URDU (Urdu) 
 & \multicolumn{1}{c|}{72.46 \iffalse $\pm$ \fi } 
 & \multicolumn{1}{c|}{66.82 \iffalse $\pm$ \fi } 
 & \multicolumn{1}{c|}{58.92 \iffalse $\pm$ \fi }    
 & \textbf{73.23} \\ \hline
\end{tabular}
\end{table*}
\label{table:cross-corpus_results}


\subsection{Cross-Language Evaluations}
\label{cross-language}
This section evaluates the models in the cross-language setting, where the audio and text inputs are from different languages. This setting is realistic and challenging for SER applications. We use four datasets with different languages and emotion classes: IEMOCAP (English), EMODB (German), URDU (Urdu), and EMOVO (Italian). We compare our model with the same the baseline, SAFRLM \cite{yang2022self}, and HCAM \cite{dutta2023hcam} 



We perform evaluations in the categorical emotions classification, where we use the four basic emotions (happy, sad, angry, and neutral) as the target labels. 
We train our model on one language and test it on another language. 
Table \ref{table:cross-corpus_results} shows the results of the models on different language pairs. Our model achieves the considerably improved performance on most of the language pairs and outperforms the baseline, SAFRLM, and HCAM  by a large margin.

From Table \ref{table:cross-corpus_results}, we observe some general trends and patterns among the different language pairs and models. 

\textit{First}, IEMOCAP has the lowest accuracy among the four datasets, indicating that it is a relatively difficult dataset for speech emotion recognition. This may be due to the fact that IEMOCAP contains spontaneous and natural speech with more variations and noise than the other datasets, which are mostly acted and recorded in controlled settings \cite{busso2008iemocap}. IEMOCAP also has more speakers and longer utterances than the other datasets, making it harder to learn the consistent and discriminative features for emotion recognition. 

\textit{Second}, URDU has the highest accuracy among the four datasets, suggesting that it is a relatively easy dataset for SER. This may be due to the fact that URDU has fewer speakers and less diversity than the other languages \cite{latif2018urdu}, making it easier to learn the acoustic and linguistic features for emotion recognition. URDU also has fewer emotion classes and simpler emotions than the other datasets \cite{latif2018urdu}, making it easier to classify the emotions correctly.

\textit{Third}, MDAT performs better when trained on English and tested on other languages than vice versa. This may imply that English has more information and richness than the other languages for speech emotion recognition \cite{schuller2010cross}, and that a model trained on English can generalise better to other languages than a model trained on other languages. Moreover, our model performs better when trained on German and tested on Italian than vice versa. This may indicate that German and Italian have some similarities in their acoustic and linguistic features for emotion recognition \cite{schuller2010cross}, and that German has more information and richness than Italian for speech emotion recognition. In contrast, our model performs worse when trained on Urdu and tested on other languages than vice versa. This may suggest that Urdu has less information and richness than the other languages for speech emotion recognition \cite{latif2018urdu}, and that a model trained on Urdu cannot generalise well to other languages than a model trained on other languages.

\textit{Fourth}, MDAT outperforms the baseline models by a large margin on most of the language pairs. This may demonstrate that our model can effectively fuse multimodal data in a progressive and interpretable way using co-attention and graph attention networks. Our model can also leverage the powerful pre-trained models for speech and text modalities: wav2vec 2.0's XLS-R and RoBERTa. Our model can capture complex dependencies between modalities, preserve modality-specific information, and enhance cross-modality and cross-language interactions. The baseline models, on the other hand, may suffer from some limitations, such as losing modality-specific information, ignoring complex dependencies between modalities, or requiring large computation and parameters. For example, the BLSTM model with simple fusion may not be able to capture the long-term dependencies and the interactions between modalities. The SAFRLM model may struggle with aligned text-audio sequences and require large computation and parameters \cite{yang2022self}. The HCAM model may ignore the global dependencies among the features and require large computation and parameters \cite{dutta2023hcam}.

\subsection{K-shot Adaptation}

\begin{table*}[]
\centering
\captionsetup{justification=centering}
\caption{Cross-language Emotion Recognition unweighted accuracy (\%) of the Bidirectional LSTM base model and MDAT model with different numbers of target samples (K-Shots) for four datasets in different languages.}
\setlength{\tabcolsep}{5pt}
\renewcommand{\arraystretch}{1.4}
\begin{tabular}{|l|l|cccccccc|}
\hline
\multirow{3}{*}{\textbf{Source}} & \multirow{3}{*}{\textbf{Target}} & \multicolumn{8}{c|}{\textbf{K-Shot UA (\%)}}                                                                                                                                                                                                                                                                       \\ \cline{3-10} 
                                 &                                  & \multicolumn{2}{c|}{\textbf{0 Shot}}                                            & \multicolumn{2}{c|}{\textbf{5 Shot}}                                            & \multicolumn{2}{c|}{\textbf{10 Shot}}                                           & \multicolumn{2}{c|}{\textbf{15 Shot}}                      \\ \cline{3-10} 
                                 &                                  & \multicolumn{1}{c|}{\textbf{Baseline}} & \multicolumn{1}{c|}{\textbf{MDAT}} & \multicolumn{1}{c|}{\textbf{Baseline}} & \multicolumn{1}{c|}{\textbf{MDAT}} & \multicolumn{1}{c|}{\textbf{Baseline}} & \multicolumn{1}{c|}{\textbf{MDAT}} & \multicolumn{1}{c|}{\textbf{Baseline}} & \textbf{MDAT} \\ \hline
\multirow{3}{*}{IEMOCAP}         & EMODB                            & \multicolumn{1}{c|}{40.78}             & \multicolumn{1}{c|}{48.48}             & \multicolumn{1}{c|}{73.18}             & \multicolumn{1}{c|}{73.70}             & \multicolumn{1}{c|}{78.77}             & \multicolumn{1}{c|}{83.80}             & \multicolumn{1}{c|}{78.98}             & 91.48             \\ \cline{2-10} 
                                 & URDU                             & \multicolumn{1}{c|}{61.18}             & \multicolumn{1}{c|}{64.43}             & \multicolumn{1}{c|}{64.44}             & \multicolumn{1}{c|}{65.22}             & \multicolumn{1}{c|}{70.68}             & \multicolumn{1}{c|}{73.86}             & \multicolumn{1}{c|}{73.19}             & 77.54             \\ \cline{2-10} 
                                 & EMOVO                            & \multicolumn{1}{c|}{73.46}             & \multicolumn{1}{c|}{85.51}             & \multicolumn{1}{c|}{77.86}             & \multicolumn{1}{c|}{85.10}             & \multicolumn{1}{c|}{78.62}             & \multicolumn{1}{c|}{87.77}             & \multicolumn{1}{c|}{83.93}             & 92.05             \\ \hline
\multirow{3}{*}{EMODB}           & IEMOCAP                          & \multicolumn{1}{c|}{49.23}             & \multicolumn{1}{c|}{55.55}             & \multicolumn{1}{c|}{48.80}             & \multicolumn{1}{c|}{51.35}             & \multicolumn{1}{c|}{51.39}             & \multicolumn{1}{c|}{59.61}             & \multicolumn{1}{c|}{52.11}             & 63.48             \\ \cline{2-10} 
                                 & URDU                             & \multicolumn{1}{c|}{72.46}             & \multicolumn{1}{c|}{74.23}             & \multicolumn{1}{c|}{72.41}             & \multicolumn{1}{c|}{76.64}             & \multicolumn{1}{c|}{75.30}             & \multicolumn{1}{c|}{83.12}             & \multicolumn{1}{c|}{78.40}             & 84.57             \\ \cline{2-10} 
                                 & EMOVO                            & \multicolumn{1}{c|}{50.54}             & \multicolumn{1}{c|}{56.25}             & \multicolumn{1}{c|}{69.99}             & \multicolumn{1}{c|}{64.18}             & \multicolumn{1}{c|}{74.01}             & \multicolumn{1}{c|}{89.36}             & \multicolumn{1}{c|}{78.40}             & 93.56             \\ \hline
\multirow{3}{*}{URDU}            & IEMOCAP                          & \multicolumn{1}{c|}{47.96}             & \multicolumn{1}{c|}{58.32}             & \multicolumn{1}{c|}{45.16}             & \multicolumn{1}{c|}{42.79}             & \multicolumn{1}{c|}{48.61}             & \multicolumn{1}{c|}{58.91}             & \multicolumn{1}{c|}{46.67}             & 59.52             \\ \cline{2-10} 
                                 & EMODB                            & \multicolumn{1}{c|}{65.35}             & \multicolumn{1}{c|}{75.32}             & \multicolumn{1}{c|}{73.18}             & \multicolumn{1}{c|}{75.42}             & \multicolumn{1}{c|}{75.42}             & \multicolumn{1}{c|}{76.49}             & \multicolumn{1}{c|}{76.51}             & 80.67             \\ \cline{2-10} 
                                 & EMOVO                            & \multicolumn{1}{c|}{65.35}             & \multicolumn{1}{c|}{76.21}             & \multicolumn{1}{c|}{65.42}             & \multicolumn{1}{c|}{77.36}             & \multicolumn{1}{c|}{69.95}             & \multicolumn{1}{c|}{83.24}             & \multicolumn{1}{c|}{70.83}             & 88.64             \\ \hline
\multirow{3}{*}{EMOVO}           & IEMOCAP                          & \multicolumn{1}{c|}{57.25}             & \multicolumn{1}{c|}{59.96}             & \multicolumn{1}{c|}{60.14}             & \multicolumn{1}{c|}{64.71}             & \multicolumn{1}{c|}{62.38}             & \multicolumn{1}{c|}{64.81}             & \multicolumn{1}{c|}{68.74}             & 69.09             \\ \cline{2-10} 
                                 & EMODB                            & \multicolumn{1}{c|}{65.60}             & \multicolumn{1}{c|}{81.60}             & \multicolumn{1}{c|}{79.43}             & \multicolumn{1}{c|}{82.81}             & \multicolumn{1}{c|}{80.73}             & \multicolumn{1}{c|}{85.41}             & \multicolumn{1}{c|}{84.62}             & 94.23             \\ \cline{2-10} 
                                 & URDU                             & \multicolumn{1}{c|}{72.46}             & \multicolumn{1}{c|}{73.23}             & \multicolumn{1}{c|}{70.12}             & \multicolumn{1}{c|}{73.50}             & \multicolumn{1}{c|}{78.76}             & \multicolumn{1}{c|}{80.45}             & \multicolumn{1}{c|}{81.16}             & 82.05             \\ \hline
\end{tabular}
\label{table:k-shot_complete}
\end{table*}

In this experiment, we evaluate our MDAT model on the k-shot adaptation experiments, where we use varying numbers of shots from 0 to 15. The shots are the number of labelled samples from the target language that are used to fine-tune the models. We compare our model with the baseline model on different language pairs and measure the accuracy as the performance metric. 
Table \ref{table:k-shot_complete} shows the results of the proposed model and the baseline model respectively.

\begin{figure}[!ht]
\centering
\includegraphics[trim=0cm 0cm 0cm 0cm, clip=true, width=1\linewidth]{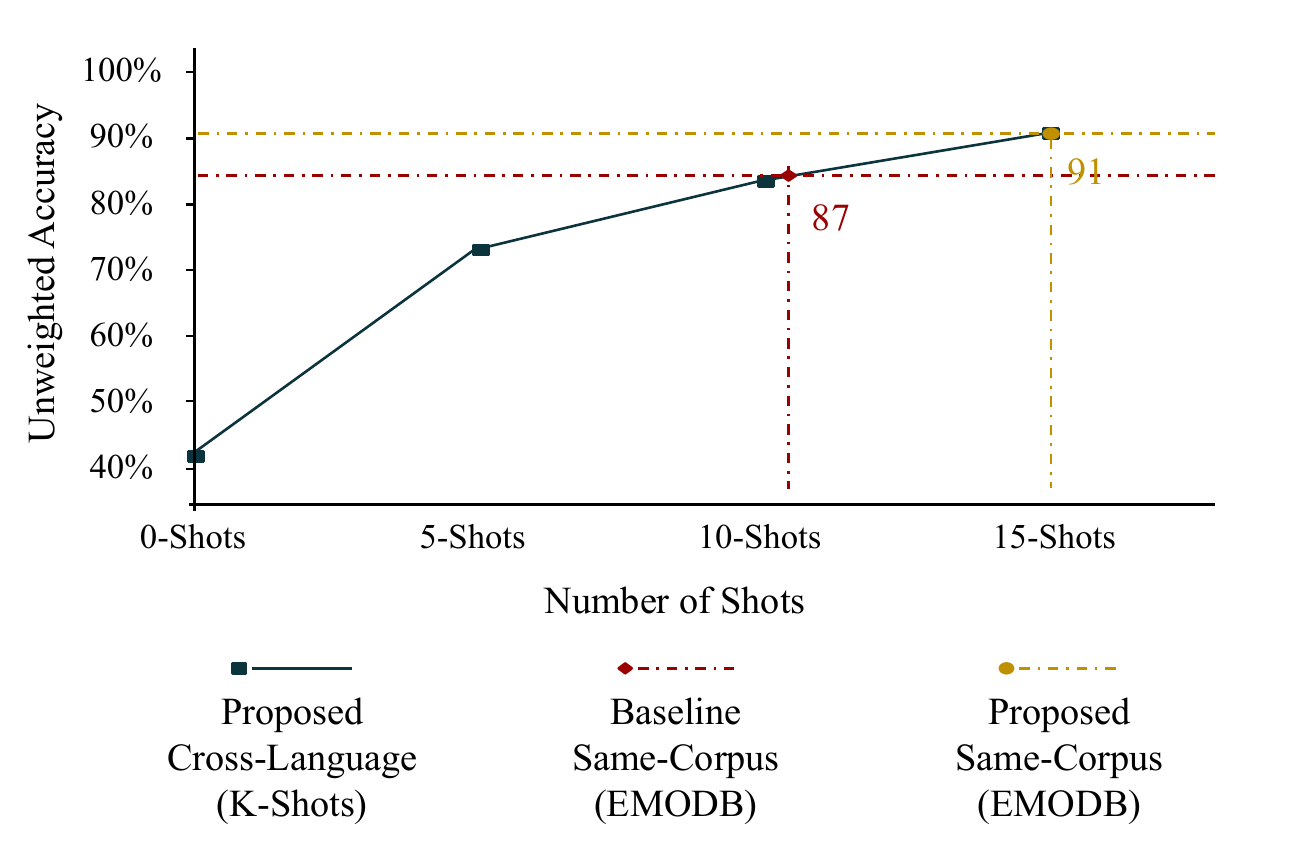}
\caption{Cross-language K-shot results: IEMOCAP to EMODB.\\
\small Unweighted accuracy of speech emotion recognition from IEMOCAP to EMODB for various shot numbers. Solid line represents the MDAT's performance, while dotted lines indicate the base model and EMODB same-corpus results.}

\label{fig:IEMOCAP-to-EMODB}
\end{figure}

\begin{figure}[!ht]
\centering
\includegraphics[trim=0cm 0cm 0cm 0cm, clip=true, width=1\linewidth]{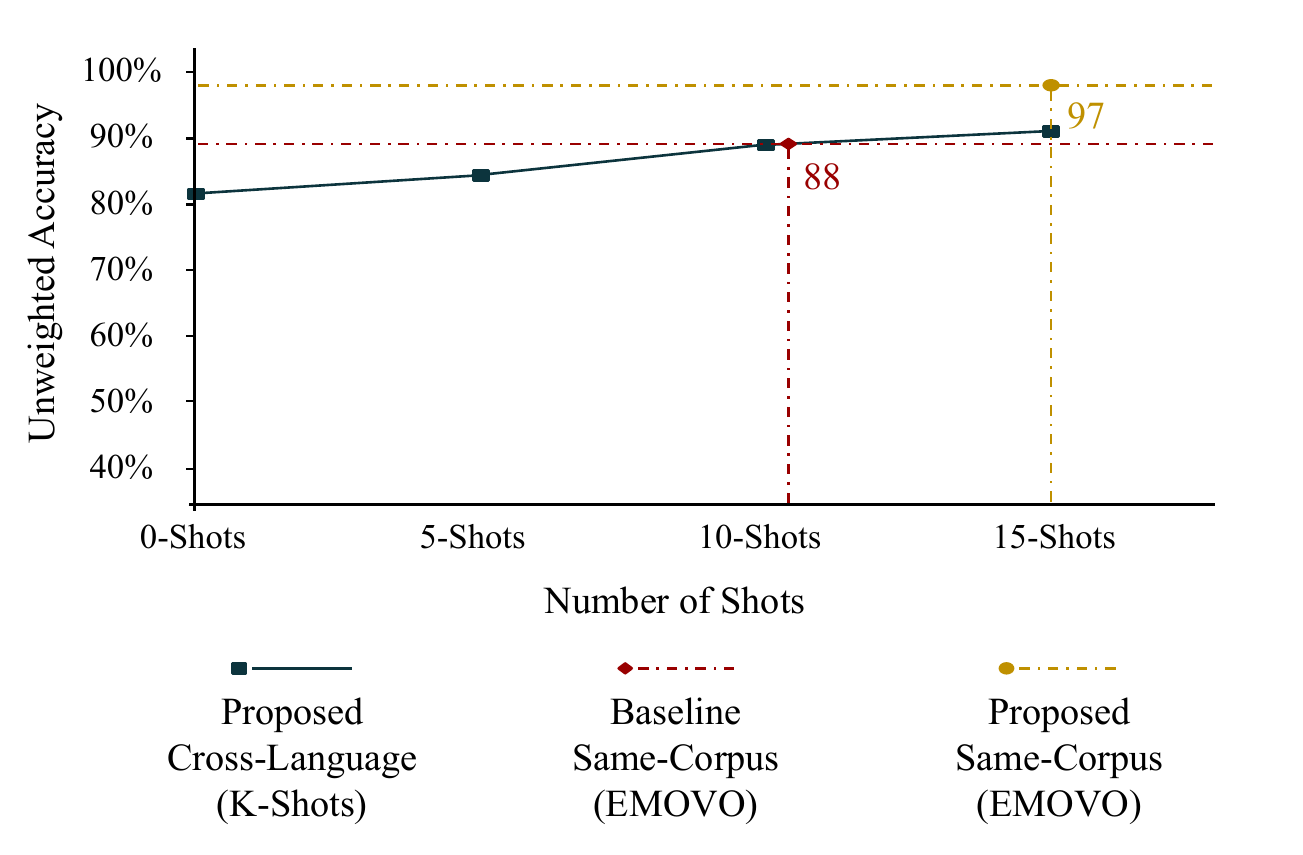}
\caption{Cross-language K-shot results: IEMOCAP to EMOVO.\\
\small Unweighted accuracy of speech emotion recognition from IEMOCAP to EMOVO for various shot numbers. Solid line represents the MDAT's performance, while dotted lines indicate the base model and EMOVO same-corpus results.}
\label{fig:IEMOCAP-to-EMOVO}
\end{figure}

We first analyse the results of our Multimodal Dual Attention Transformer (MDAT) model in Table \ref{table:k-shot_complete}. 

Our proposed model achives considerable improvements on all the language pairs when increasing the number of shots. For example, when the model is trained using English and tested on German, our model increases its accuracy from 42.48 \% with 0 shots to 91.48 \% with 15 shots. Similarly, when trained on English and tested on Italian, our model increases its accuracy from 82.51 \& with 0 shots to 92.05 \% with 15 shots. This shows that our model can effectively adapt to a new low-resource language with a small amount of labelled data. 


We then compare the results of our proposed MDAT model with the baseline model in Table \ref{table:k-shot_complete}. We can observe that our proposed model outperforms the baseline model on all language pairs and shots, demonstrating its superiority and generalisation for cross-language emotion recognition. We can also observe that our proposed model surpasses the baseline model by a large margin on some language pairs and shots. For example, when trained using German and tested on Italian with 15 shots, our proposed model outperforms the baseline model by more than 9 \%. Similarly, when the models are trained using Urdu and tested on English with 15 shots, our proposed model outperforms the baseline model by more than 12 \% points.

We attribute these results to the advantages of our proposed model over the baseline model in terms of feature extraction and dual attention mechanisms.

\textit{First}, our proposed model leverages multilingual pre-trained models for each modality, namely wav2vec 2.0 \cite{baevski2020wav2vec} for the audio modality and RoBERTa \cite{liu2019roberta} for the text modality. These models are pre-trained on large-scale multilingual corpora and can capture rich and universal acoustic and linguistic features for different languages and domains \cite{2108.09669}. These features help our model to generalise better to new languages with a small amount of labelled data. In contrast, the baseline model encodes the audio and text inputs separately using a bidirectional LSTM \cite{hochreiter1997long}, which may not be able to capture the nuances and variations of different languages and domains. These features limit the generalisation ability of the baseline model to new languages with a small amount of labelled data.

\textit{Second}, our proposed model utilise the co-attention and graph attention layers to learn emotionally salient features for cross-language SER. The co-attention network aligns the audio and text features at the local level and generates a multimodal context vector for each feature. The graph attention network aggregates the multimodal context vectors at the global level and generates a multimodal representation vector for each input. These networks capture complex dependencies between modalities and effectively preserve modality-specific information. These information help our model to exploit the complementary and supplementary aspects of different modalities for emotion recognition. In contrast, the baseline model fuses the audio and text features at the end of the encoding process using a simple concatenation technique \cite{chen2019complementary}. This technique may not be able to capture the complex dependencies between modalities and may lose modality-specific information. These information limit the exploitation ability of the baseline model to use different modalities for emotion recognition.

\begin{table*}[!t]
\centering
\captionsetup{justification=centering}
\caption{Results from the Ablation experiments evaluating the performance of each module of the MDAT model, using unweighted accuracy as the assessment metric.}
\setlength{\tabcolsep}{5pt}
\renewcommand{\arraystretch}{1.4}
\begin{tabular}{|>{\centering\arraybackslash}m{1.0cm}|>{\centering\arraybackslash}m{6cm}|c|c|c|ccc|}
\hline
\multirow{3}{*}{\textbf{Model}} & \multirow{3}{*}{\textbf{Configuration}} & \multirow{3}{*}{\textbf{\begin{tabular}[c]{@{}c@{}}Graph\\ Attention\end{tabular}}} & \multirow{3}{*}{\textbf{\begin{tabular}[c]{@{}c@{}}Co-\\ Attention\end{tabular}}} & \multirow{3}{*}{\textbf{\begin{tabular}[c]{@{}c@{}}Transformer\\ Encoder\end{tabular}}} & \multicolumn{3}{c|}{\textbf{Cross-Language UA (\%)}} \\ \cline{6-8} 
 & & & & & \multicolumn{3}{c|}{\textbf{IEMOCAP (English) to}} \\ 
 & & & & & \multicolumn{1}{c|}{\textbf{\begin{tabular}[c]{@{}c@{}}EMODB\\ (German)\end{tabular}}} & \multicolumn{1}{c|}{\textbf{\begin{tabular}[c]{@{}c@{}}EMOVO\\ (Italian)\end{tabular}}} & \textbf{\begin{tabular}[c]{@{}c@{}}URDU\\ (Urdu)\end{tabular}} \\ \hline
\multirow[b]{2}{=}{\centering 1} & \includegraphics[width=5.5cm]{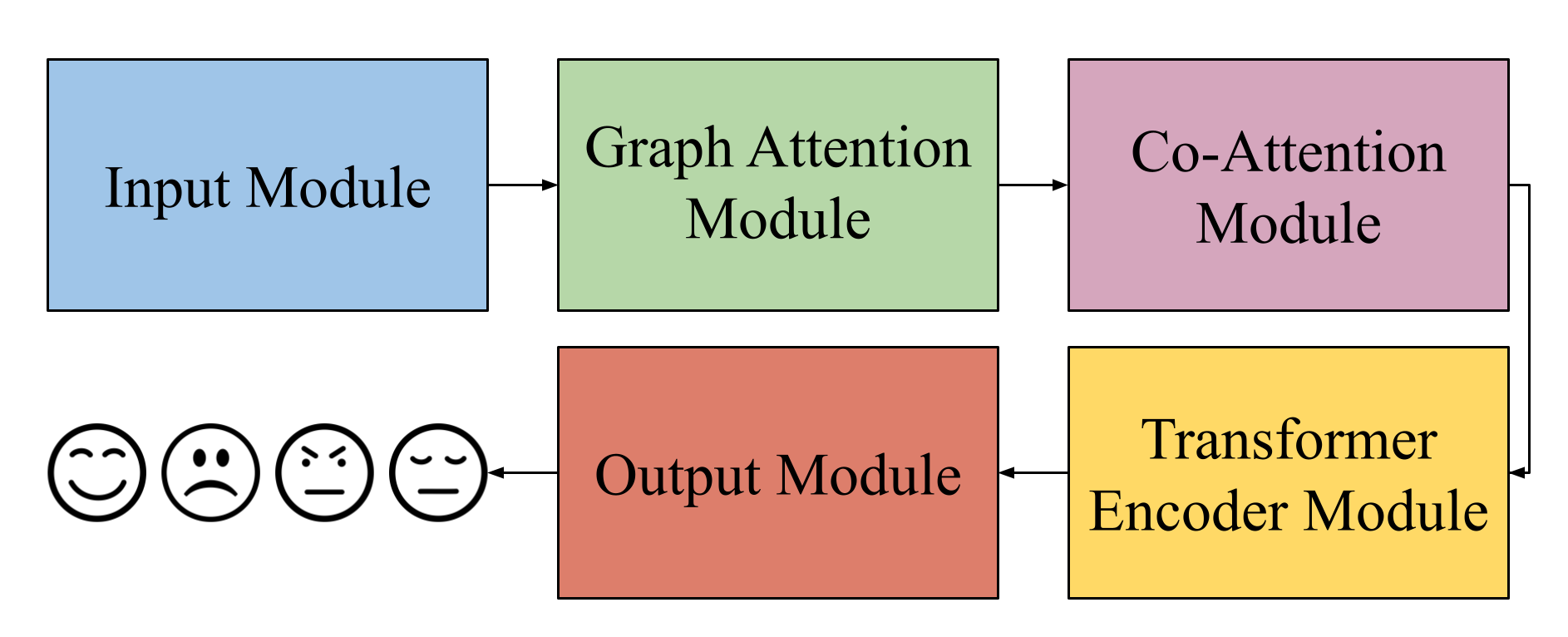} & $\cmark{}$ & $\cmark{}$ & $\cmark{}$ & \multicolumn{1}{c|}{42.48} & \multicolumn{1}{c|}{85.51} & 64.43 \\ \hline

\multirow[b]{2}{=}{\centering 2} & \includegraphics[width=5.5cm]{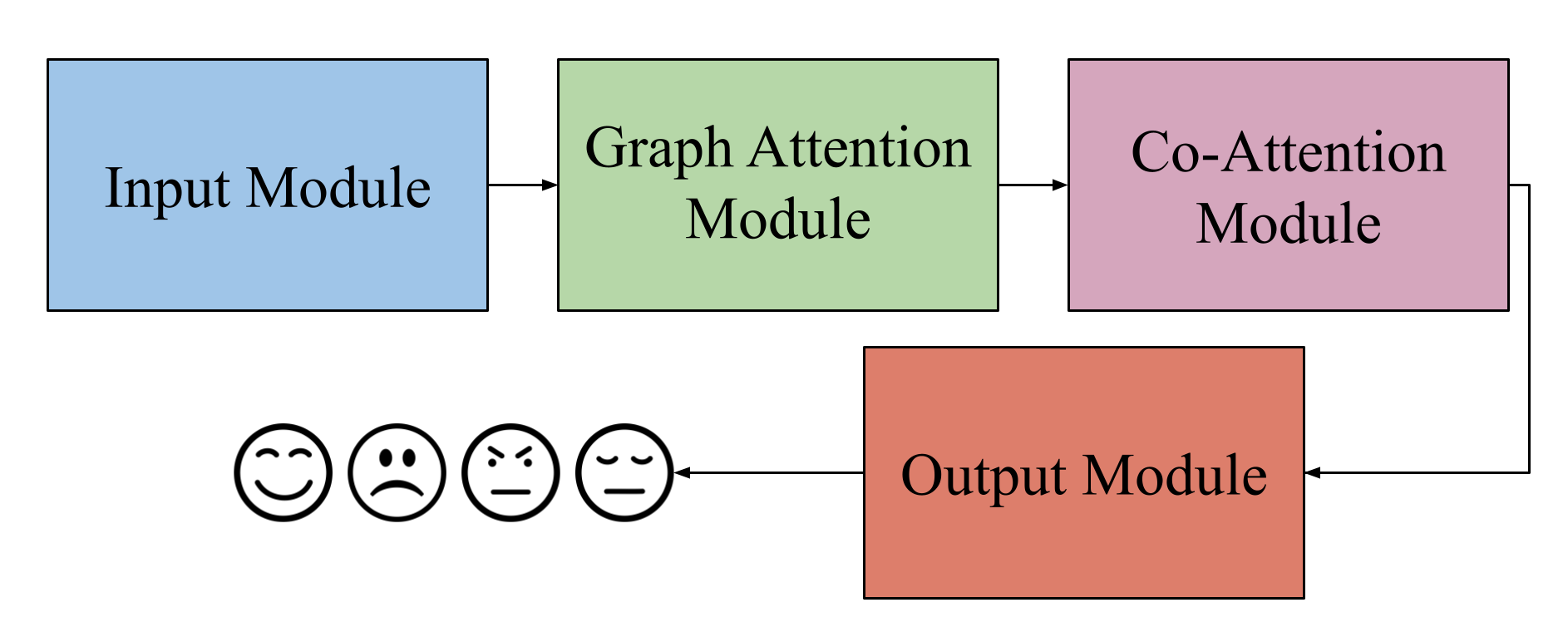} & $\cmark{}$ & $\cmark{}$ & $\xmark{}$ & \multicolumn{1}{c|}{39.21} & \multicolumn{1}{c|}{80.68} & 56.52 \\ \hline

\multirow[b]{2}{=}{\centering 3} & \includegraphics[width=5.5cm]{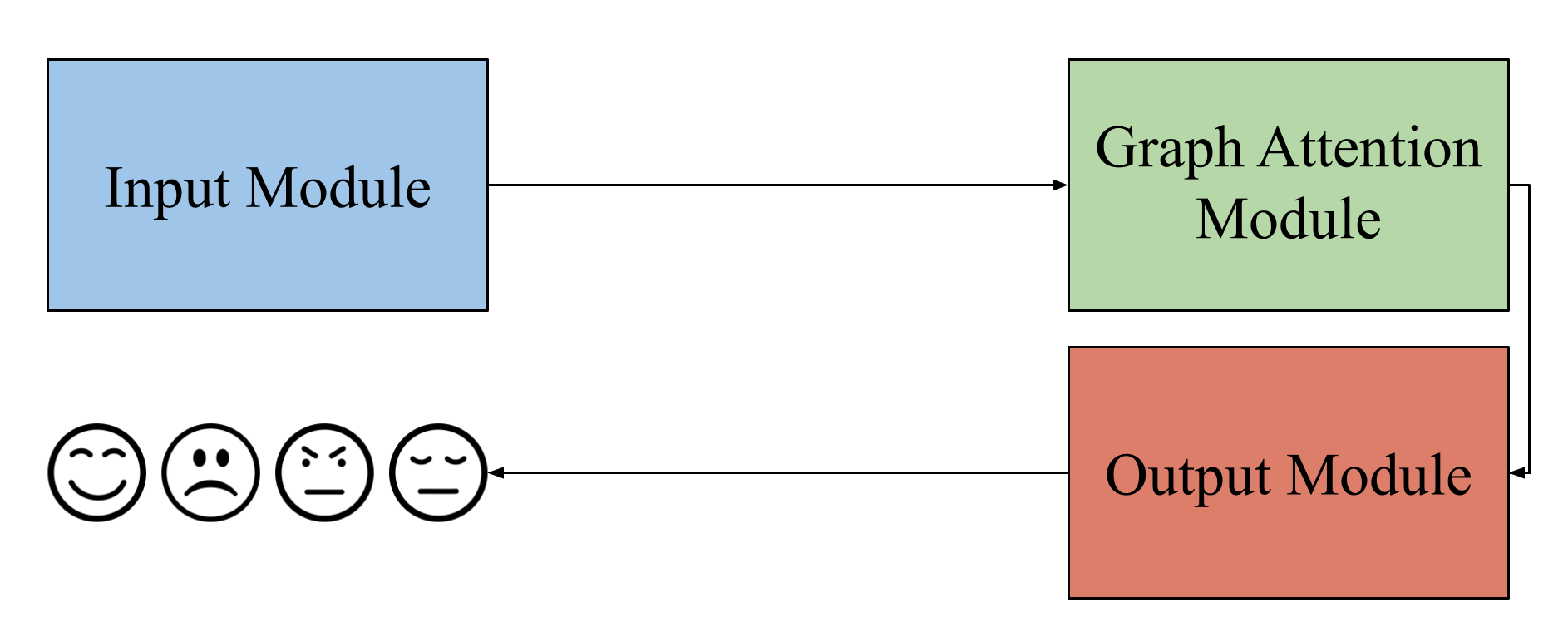} & $\cmark{}$ & $\xmark{}$ & $\xmark{}$ & \multicolumn{1}{c|}{41.58} & \multicolumn{1}{c|}{76.13} & 45.41 \\ \hline

\multirow[b]{2}{=}{\centering 4} & \includegraphics[width=5.5cm]{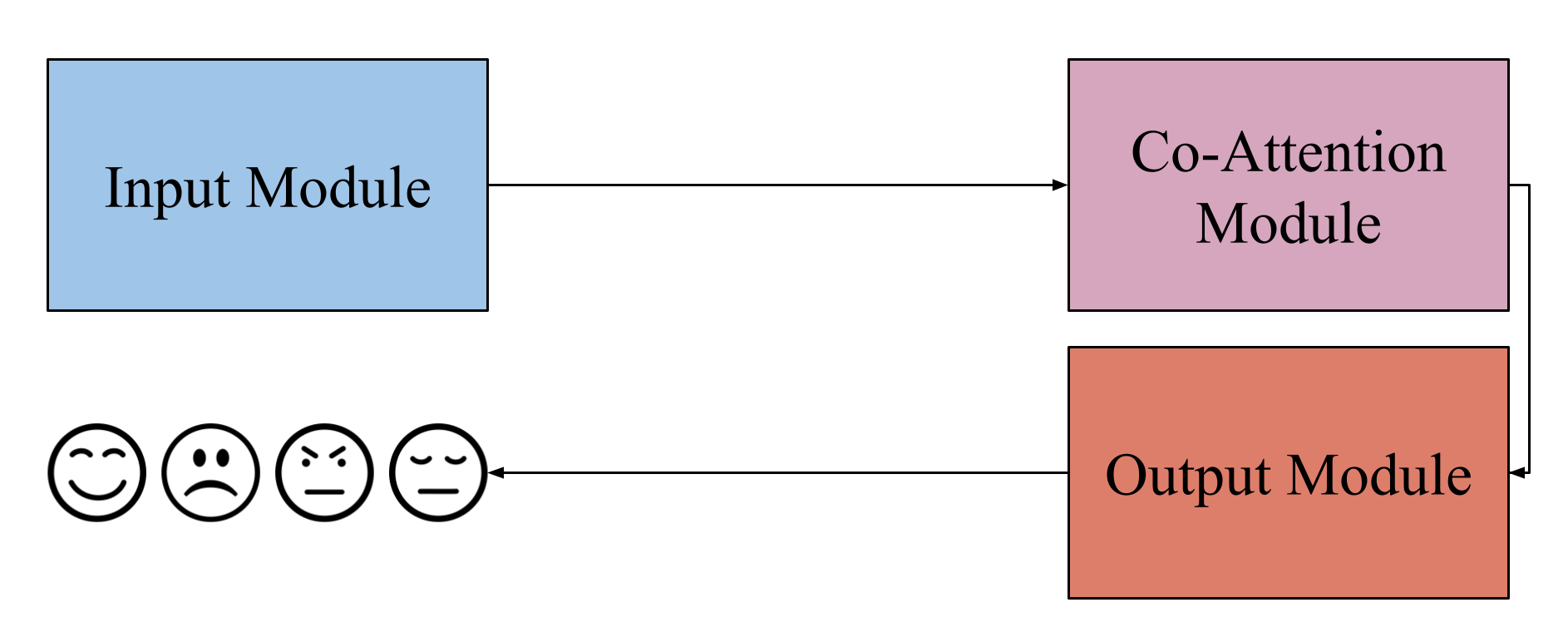} & $\xmark{}$ & $\cmark{}$ & $\xmark{}$ & \multicolumn{1}{c|}{39.06} & \multicolumn{1}{c|}{41.82} & 53.14 \\ \hline

\multirow[b]{2}{=}{\centering 5} & \includegraphics[width=5.5cm]{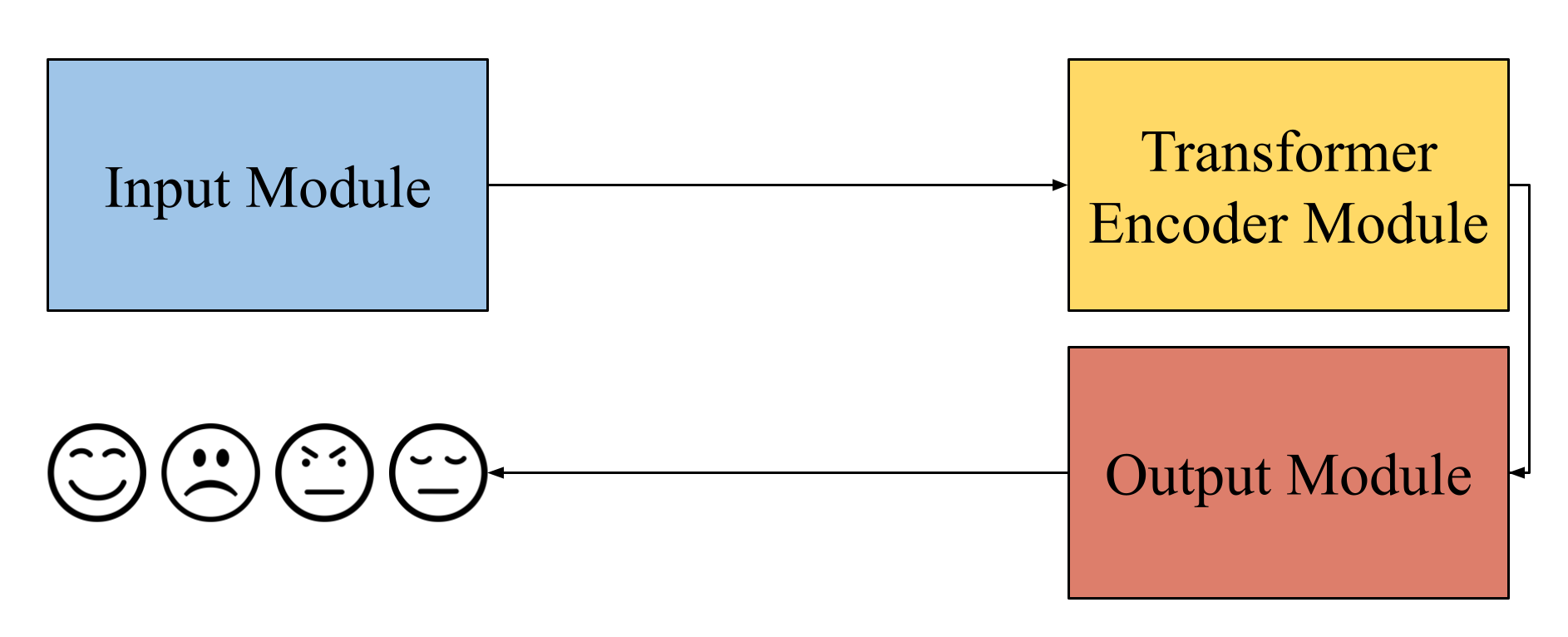} & $\xmark{}$ & $\xmark{}$ & $\cmark{}$ & \multicolumn{1}{c|}{34.80} & \multicolumn{1}{c|}{46.59} & 36.23 \\ \hline

\multirow[b]{2}{=}{\centering 6} & \includegraphics[width=5.5cm]{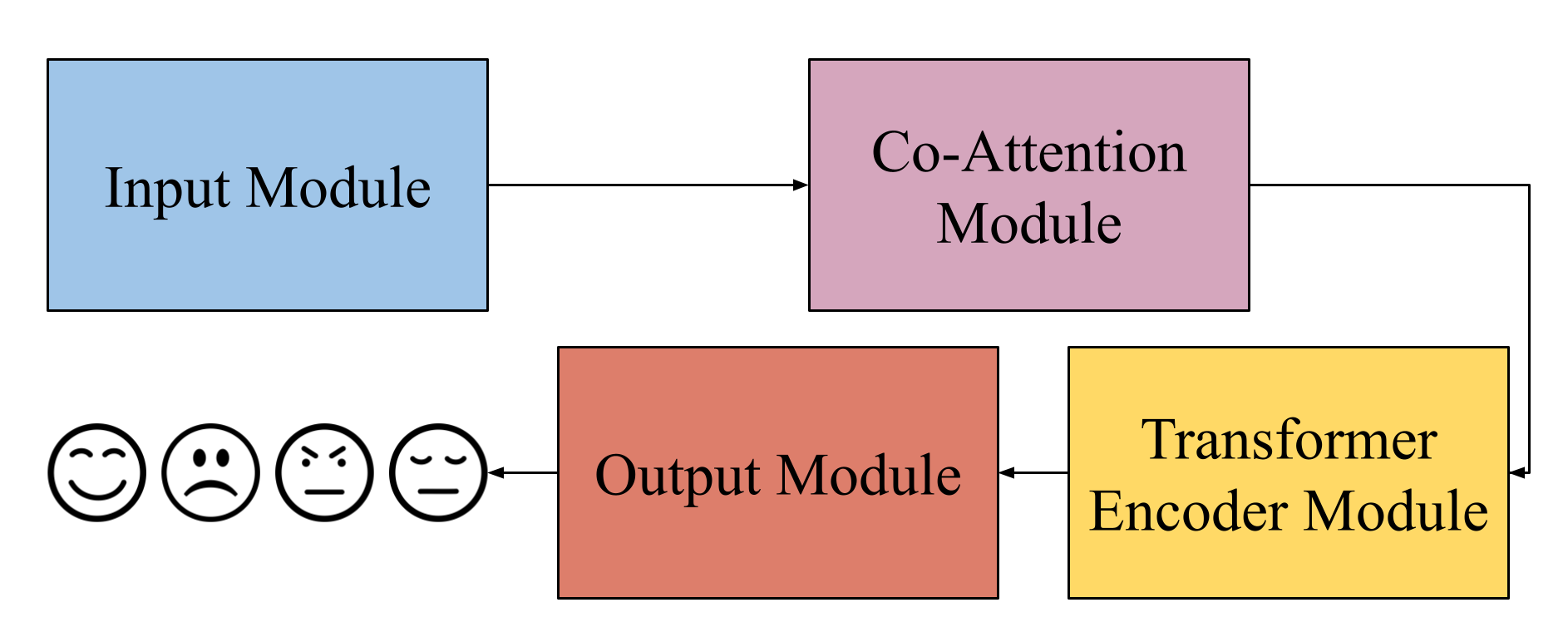} & $\xmark{}$ & $\cmark{}$ & $\cmark{}$ & \multicolumn{1}{c|}{38.72} & \multicolumn{1}{c|}{57.95} & 36.71 \\ \hline

\multirow[b]{2}{=}{\centering 7} & \includegraphics[width=5.5cm]{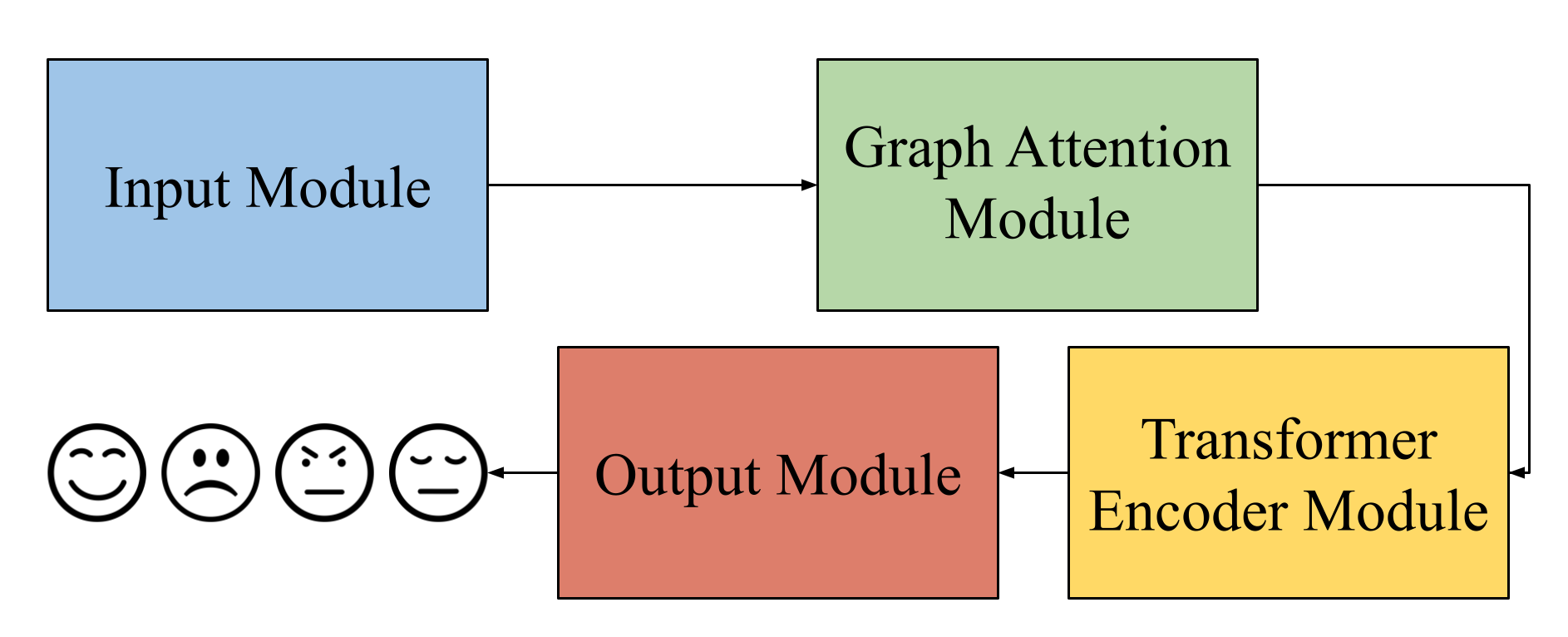} & $\cmark{}$ & $\xmark{}$ & $\cmark{}$ & \multicolumn{1}{c|}{39.70} & \multicolumn{1}{c|}{30.91} & 42.99 \\ \hline

\end{tabular}
\label{table:ablation}

\end{table*}

\textit{Third}, in contrast to the baseline, our proposed model enhances the multimodal SER using a transformer encoder layer \cite{vaswani2017attention}. The transformer encoder layer consists of a multi-head self-attention sublayer and a feed-forward sublayer. This layer enhances cross-modality and cross-language interactions and generates more expressive and robust multimodal features for emotion recognition \cite{huang2020multimodal}. These features help our model to handle different emotions and scenarios in different languages and domains.

\subsection{Ablation Experiments}
\label{sec:ablation}

The MDAT model has three main modules: \textit{graph attention}, which applies attention mechanisms to graph-structured data; \textit{co-attention}, which computes the attention weights for each modality based on the other modality; and \textit{transformer encoder}, which uses self-attention and feed-forward layers to encode one modality. In this section, we present the results of an ablation study to test the contribution and impact of each module in our proposed MDAT model for cross-language SER with different combinations of these modules and measured their performance using unweighted accuracy (Table \ref{table:ablation}). We used three cross-language scenarios: IEMOCAP (English) to EMODB (German); IEMOCAP (English) to EMOVO (Italian); and IEMOCAP (English) to URDU (Urdu). Our results in Table \ref{table:ablation} indicate that graph attention and co-attention are essential for cross-language emotion recognition, and that transformer encoder can enhance the performance when combined with them.

The complete MDAT model (Model 1) achieves the highest unweighted accuracy in all three scenarios, demonstrating the benefits of incorporating all three modules. This is because each module leverages complementary information: graph attention captures consistent syntactic dependencies, co-attention aligns different modalities, and transformer encoder encodes self-attention and feed-forward features. Removing any of the modules from the complete proposed model (Model 2 and Model 7) leads to a decrease in unweighted accuracy across all scenarios. This indicates that each module contributes significantly to the model's performance. Graph attention provides structural information for modeling long-range dependencies between words, co-attention reduces noise and redundancy in modalities, and transformer encoder enhances the encoding of one modality.

Among the modules, graph attention emerges as the most crucial for the task (Model 3 vs. Model 4, Model 5 vs. Model 6, and Model 7 vs. Model 6). Its addition to any configuration consistently improves unweighted accuracy, highlighting its ability to leverage universal dependency structure of languages, which is more consistent and transferable across languages than other features. Graph attention also helps to model long-range dependencies between words that are not directly connected in the dependency tree, which may be crucial for emotion recognition.

Co-attention ranks as the second most important module, improving unweighted accuracy in all configurations 
except for one specific case (Model 3 vs. Model 2). Co-attention's alignment of modalities and reduction of noise and redundancy are valuable, although it may not be sufficient without the structural information provided by graph attention. Similar to other modules, transformer encoder layer is playing its role towards achieving improved results for cross-language SER. It can be noted that transformer encoder layer is when it is combined with co-attention (Model 6 vs. Model 4). This shows all the MDAT's modules are chosen carefully to achieve improve results for cross-language SER.


\section{Conclusions and Outlook}
\label{sec:Conclusions}

In this paper, we proposed a novel multimodal framework to improve cross-language speech emotion recognition (SER) using dual attention layers including co-attention and graph attention. Our technique can effectively leverage the powerful pre-trained models including wav2vec 2.0 and RoBERTa for extracting multi-language speech and text embedding. We evaluated our technique on four datasets with different languages and emotion classes: IEMOCAP (English), EMODB (German), URDU (Urdu), and EMOVO (Italian). We compared our proposed model, Multimodal Dual Attention Transformer (MDAT) with baseline and recent multimodal frameworks including SAFRLM with self-adjusting fusion, and HCAM with hierarchical cross-attention on categorical cross-language SER. The results showed that our proposed framework achieved considerably improved results by capturing complex dependencies between modalities and preserving modality-specific information



Our work leads to the following overarching conclusions in the realm of cross-language SER: (1) Cross-language SER is a challenging yet highly significant task with diverse practical applications, such as multilingual virtual assistants, cross-cultural communication, and emotion-aware machine translation. (2) The incorporation of multimodal learning serves as a crucial technique for cross-language SER, effectively harnessing the complementary and redundant information present in different modalities and languages. Additionally, (3) leveraging pre-trained models proves to be instrumental in extracting multi-language embeddings from speech and text modalities, enabling the encoding of multimodal input data into generalized representations that enhance the overall performance of multi-modal cross-language SER. Notably, (4) co-attention and graph attention networks emerge as effective mechanisms for facilitating multimodal learning by effectively capturing complex dependencies among speech samples from different languages, thereby improving the accuracy of cross-language SER. Finally, (5) k-shot adaptation techniques demonstrate their usefulness in addressing the scarcity of labeled data in low-resource languages, as they successfully enhance model performance even with a limited number of labeled samples. These findings collectively contribute to the advancement of cross-language SER, paving the way for future research and practical applications.



This work opens several opportunities for future work in the field of cross-language SER. These include the exploration of the visual modality to further enhance the performance of MDAT. Furthermore, it is important to evaluate the effectiveness of multimodal frameworks in more realistic and diverse scenarios, such as noisy environments, in the face of adversarial attacks, and with spontaneous speech and mixed emotions.


\begin{thebibliography}{100}
\providecommand{\url}[1]{#1}
\csname url@samestyle\endcsname
\providecommand{\newblock}{\relax}
\providecommand{\bibinfo}[2]{#2}
\providecommand{\BIBentrySTDinterwordspacing}{\spaceskip=0pt\relax}
\providecommand{\BIBentryALTinterwordstretchfactor}{4}
\providecommand{\BIBentryALTinterwordspacing}{\spaceskip=\fontdimen2\font plus
\BIBentryALTinterwordstretchfactor\fontdimen3\font minus
  \fontdimen4\font\relax}
\providecommand{\BIBforeignlanguage}[2]{{%
\expandafter\ifx\csname l@#1\endcsname\relax
\typeout{** WARNING: IEEEtran.bst: No hyphenation pattern has been}%
\typeout{** loaded for the language `#1'. Using the pattern for}%
\typeout{** the default language instead.}%
\else
\language=\csname l@#1\endcsname
\fi
#2}}
\providecommand{\BIBdecl}{\relax}
\BIBdecl

\bibitem{latif2020speech}
S.~Latif, J.~Qadir, A.~Qayyum, M.~Usama, and S.~Younis, ``Speech technology for
  healthcare: Opportunities, challenges, and state of the art,'' \emph{IEEE
  Reviews in Biomedical Engineering}, vol.~14, pp. 342--356, 2020.

\bibitem{rana2019automated}
R.~Rana, S.~Latif, R.~Gururajan, A.~Gray, G.~Mackenzie, G.~Humphris, and
  J.~Dunn, ``Automated screening for distress: A perspective for the future,''
  \emph{European journal of cancer care}, vol.~28, no.~4, p. e13033, 2019.

\bibitem{zepf2020driver}
S.~Zepf, J.~Hernandez, A.~Schmitt, W.~Minker, and R.~W. Picard, ``Driver
  emotion recognition for intelligent vehicles: A survey,'' \emph{ACM Computing
  Surveys (CSUR)}, vol.~53, no.~3, pp. 1--30, 2020.

\bibitem{tavi2020prosodic}
L.~Tavi, ``Prosodic cues of speech under stress: phonetic exploration of
  finnish emergency calls,'' Ph.D. dissertation, It{\"a}-Suomen yliopisto,
  2020.

\bibitem{yadegaridehkordi2019affective}
E.~Yadegaridehkordi, N.~F. B.~M. Noor, M.~N.~B. Ayub, H.~B. Affal, and N.~B.
  Hussin, ``Affective computing in education: A systematic review and future
  research,'' \emph{Computers \& Education}, vol. 142, p. 103649, 2019.

\bibitem{vanderplaetse2020improved}
B.~Vanderplaetse and S.~Dupont, ``Improved soccer action spotting using both
  audio and video streams,'' in \emph{Proceedings of the IEEE/CVF Conference on
  Computer Vision and Pattern Recognition Workshops}, 2020, pp. 896--897.

\bibitem{aldeneh2021you}
Z.~Aldeneh and E.~M. Provost, ``You're not you when you're angry: Robust
  emotion features emerge by recognizing speakers,'' \emph{IEEE Transactions on
  Affective Computing}, 2021.

\bibitem{nediyanchath2020multi}
A.~Nediyanchath, P.~Paramasivam, and P.~Yenigalla, ``Multi-head attention for
  speech emotion recognition with auxiliary learning of gender recognition,''
  in \emph{ICASSP 2020-2020 IEEE International Conference on Acoustics, Speech
  and Signal Processing (ICASSP)}.\hskip 1em plus 0.5em minus 0.4em\relax IEEE,
  2020, pp. 7179--7183.

\bibitem{wang2017learning}
Z.-Q. Wang and I.~Tashev, ``Learning utterance-level representations for speech
  emotion and age/gender recognition using deep neural networks,'' in
  \emph{2017 IEEE international conference on acoustics, speech and signal
  processing (ICASSP)}.\hskip 1em plus 0.5em minus 0.4em\relax IEEE, 2017, pp.
  5150--5154.

\bibitem{latif2018cross}
S.~Latif, A.~Qayyum, M.~Usman, and J.~Qadir, ``Cross lingual speech emotion
  recognition: Urdu vs. western languages,'' in \emph{2018 International
  Conference on Frontiers of Information Technology (FIT)}.\hskip 1em plus
  0.5em minus 0.4em\relax IEEE, 2018, pp. 88--93.

\bibitem{laukka2014evidence}
P.~Laukka, D.~Neiberg, and H.~A. Elfenbein, ``Evidence for cultural dialects in
  vocal emotion expression: acoustic classification within and across five
  nations.'' \emph{Emotion}, vol.~14, no.~3, p. 445, 2014.

\bibitem{hinton2006fast}
G.~E. Hinton, S.~Osindero, and Y.-W. Teh, ``A fast learning algorithm for deep
  belief nets,'' \emph{Neural computation}, vol.~18, no.~7, pp. 1527--1554,
  2006.

\bibitem{lecun1989handwritten}
Y.~LeCun, B.~Boser, J.~Denker, D.~Henderson, R.~Howard, W.~Hubbard, and
  L.~Jackel, ``Handwritten digit recognition with a back-propagation network,''
  \emph{Advances in neural information processing systems}, vol.~2, 1989.

\bibitem{latif2020deep}
S.~Latif, R.~Rana, S.~Khalifa, R.~Jurdak, and B.~W. Schuller, ``Deep
  architecture enhancing robustness to noise, adversarial attacks, and
  cross-corpus setting for speech emotion recognition,'' \emph{Proc.
  Interspeech 2020}, pp. 2327--2331, 2020.

\bibitem{neumann2017attentive}
M.~Neumann and N.~T. Vu, ``Attentive convolutional neural network based speech
  emotion recognition: A study on the impact of input features, signal length,
  and acted speech,'' \emph{Proc. Interspeech 2017}, pp. 1263--1267, 2017.

\bibitem{atila2021attention}
O.~Atila and A.~{\c{S}}eng{\"u}r, ``Attention guided 3d cnn-lstm model for
  accurate speech based emotion recognition,'' \emph{Applied Acoustics}, vol.
  182, p. 108260, 2021.

\bibitem{latif2019direct}
\BIBentryALTinterwordspacing
S.~Latif, R.~Rana, S.~Khalifa, R.~Jurdak, and J.~Epps, ``{Direct Modelling of
  Speech Emotion from Raw Speech},'' in \emph{Proc. Interspeech 2019}, 2019,
  pp. 3920--3924. [Online]. Available:
  \url{http://dx.doi.org/10.21437/Interspeech.2019-3252}
\BIBentrySTDinterwordspacing

\bibitem{schaefer2008learning}
A.~M. Schaefer, S.~Udluft, and H.-G. Zimmermann, ``Learning long-term
  dependencies with recurrent neural networks,'' \emph{Neurocomputing},
  vol.~71, no. 13-15, p. 2481–2488, 2008.

\bibitem{zhang2021recurrent}
J.~Zhang, Y.~Zeng, and B.~Starly, ``Recurrent neural networks with long term
  temporal dependencies in machine tool wear diagnosis and prognosis,''
  \emph{SN Applied Sciences}, vol.~3, no.~3, p. 1–14, 2021.

\bibitem{liu2020subtraction}
Y.~Liu, X.~Li, S.~Li, S.~Zhang, Z.~Liu, and X.~Zhang, ``Subtraction gates:
  Another way to learn long-term dependencies in recurrent neural networks,''
  \emph{IEEE Transactions on Neural Networks and Learning Systems}, 2020.

\bibitem{latif2023transformers}
S.~Latif, A.~Zaidi, H.~Cuayahuitl, F.~Shamshad, M.~Shoukat, and J.~Qadir,
  ``Transformers in speech processing: A survey,'' \emph{arXiv preprint
  arXiv:2303.11607}, 2023.

\bibitem{latif2021survey}
S.~Latif, R.~Rana, S.~Khalifa, R.~Jurdak, J.~Qadir, and B.~W. Schuller,
  ``Survey of deep representation learning for speech emotion recognition,''
  \emph{IEEE Transactions on Affective Computing}, 2021.

\bibitem{poria2017review}
S.~Poria, E.~Cambria, R.~Bajpai, and A.~Hussain, ``A review of affective
  computing: From unimodal analysis to multimodal fusion,'' \emph{Information
  Fusion}, vol.~37, pp. 98--125, 2017.

\bibitem{zadeh2018multimodal}
A.~B. Zadeh, P.~P. Liang, S.~Poria, E.~Cambria, and L.-P. Morency, ``Multimodal
  language analysis in the wild: Cmu-mosei dataset and interpretable dynamic
  fusion graph,'' in \emph{Proceedings of the 56th Annual Meeting of the
  Association for Computational Linguistics (Volume 1: Long Papers)}, 2018, pp.
  2236--2246.

\bibitem{devlin2019bert}
J.~Devlin, M.-W. Chang, K.~Lee, and K.~Toutanova, ``Bert: Pre-training of deep
  bidirectional transformers for language understanding,'' in \emph{Proceedings
  of the 2019 Conference of the North American Chapter of the Association for
  Computational Linguistics: Human Language Technologies (NAACL-HLT)}, 2019, p.
  4171–4186.

\bibitem{liu2019roberta}
Y.~Liu, M.~Ott, N.~Goyal, J.~Du, M.~Joshi, D.~Chen, O.~Levy, M.~Lewis,
  L.~Zettlemoyer, and V.~Stoyanov, ``{RoBERTa: A Robustly Optimized BERT
  Pretraining Approach},'' \emph{arXiv preprint arXiv:1907.11692}, 2019.

\bibitem{raffel2020exploring}
C.~Raffel, N.~Shazeer, A.~Roberts, K.~Lee, S.~Narang, M.~Matena, Y.~Zhou,
  W.~Li, and P.~J. Liu, ``Exploring the limits of transfer learning with a
  unified text-to-text transformer,'' \emph{The Journal of Machine Learning
  Research}, vol.~21, no.~1, pp. 5485--5551, 2020.

\bibitem{babu2021xls}
A.~Babu, C.~Wang, A.~Tjandra, K.~Lakhotia, Q.~Xu, N.~Goyal, K.~Singh, P.~von
  Platen, Y.~Saraf, J.~Pino \emph{et~al.}, ``Xls-r: Self-supervised
  cross-lingual speech representation learning at scale,'' \emph{arXiv preprint
  arXiv:2111.09296}, 2021.

\bibitem{velivckovic2017graph}
P.~Veli{\v{c}}kovi{\'c}, G.~Cucurull, A.~Casanova, A.~Romero, P.~Lio, and
  Y.~Bengio, ``Graph attention networks,'' \emph{arXiv preprint
  arXiv:1710.10903}, 2017.

\bibitem{lu2016hierarchical}
J.~Lu, J.~Yang, D.~Batra, and D.~Parikh, ``Hierarchical question-image
  co-attention for visual question answering,'' \emph{Advances in neural
  information processing systems}, vol.~29, 2016.

\bibitem{deng2020low}
S.~Deng, N.~Zhang, Z.~Sun, J.~Chen, and H.~Chen, ``When low resource nlp meets
  unsupervised language model: Meta-pretraining then meta-learning for few-shot
  text classification (student abstract),'' in \emph{Proceedings of the AAAI
  Conference on Artificial Intelligence}, vol.~34, no.~10, 2020, pp.
  13\,773--13\,774.

\bibitem{lahoti2022survey}
P.~Lahoti, N.~Mittal, and G.~Singh, ``A survey on nlp resources, tools, and
  techniques for marathi language processing,'' \emph{ACM Transactions on Asian
  and Low-Resource Language Information Processing}, vol.~22, no.~2, pp. 1--34,
  2022.

\bibitem{godard2017very}
P.~Godard, G.~Adda, M.~Adda-Decker, J.~Benjumea, L.~Besacier,
  J.~Cooper-Leavitt, G.-N. Kouarata, L.~Lamel, H.~Maynard, M.~M{\"u}ller
  \emph{et~al.}, ``A very low resource language speech corpus for computational
  language documentation experiments,'' \emph{arXiv preprint arXiv:1710.03501},
  2017.

\bibitem{schuller2010interspeech}
F.~Schuller, S.~Steidl, A.~Batliner, G.~Rigoll, and K.~Lang, ``Interspeech
  2010: 11th annual conference of the international speech communication
  association,'' in \emph{Interspeech}, 2010.

\bibitem{schuller2013interspeech}
B.~Schuller, S.~Steidl, A.~Batliner, A.~Vinciarelli, K.~Scherer, F.~Ringeval,
  M.~Chetouani, F.~Weninger, F.~Eyben, E.~Marchi \emph{et~al.}, ``The
  interspeech 2013 computational paralinguistics challenge: Social signals,
  conflict, emotion, autism,'' in \emph{Proceedings INTERSPEECH 2013, 14th
  Annual Conference of the International Speech Communication Association,
  Lyon, France}, 2013.

\bibitem{ozseven2019novel}
T.~{\"O}zseven, ``A novel feature selection method for speech emotion
  recognition,'' \emph{Applied Acoustics}, vol. 146, pp. 320--326, 2019.

\bibitem{ahn2021cross}
Y.~Ahn, S.~J. Lee, and J.~W. Shin, ``Cross-corpus speech emotion recognition
  based on few-shot learning and domain adaptation,'' \emph{IEEE Signal
  Processing Letters}, vol.~28, pp. 1190--1194, 2021.

\bibitem{kshirsagar2022cross}
S.~Kshirsagar and T.~H. Falk, ``Cross-language speech emotion recognition using
  bag-of-word representations, domain adaptation, and data augmentation,''
  \emph{Sensors}, vol.~22, no.~17, p. 6445, 2022.

\bibitem{latif2022multitask}
S.~Latif, R.~Rana, S.~Khalifa, R.~Jurdak, and B.~W. Schuller, ``Multitask
  learning from augmented auxiliary data for improving speech emotion
  recognition,'' \emph{IEEE Transactions on Affective Computing}, 2022.

\bibitem{shen2020wise}
G.~Shen, R.~Lai, R.~Chen, Y.~Zhang, K.~Zhang, Q.~Han, and H.~Song, ``Wise:
  Word-level interaction-based multimodal fusion for speech emotion
  recognition.'' in \emph{Interspeech}, 2020, pp. 369--373.

\bibitem{atmaja2022survey}
B.~T. Atmaja, A.~Sasou, and M.~Akagi, ``Survey on bimodal speech emotion
  recognition from acoustic and linguistic information fusion,'' \emph{Speech
  Communication}, 2022.

\bibitem{kim2021contrastive}
D.~Kim and B.~C. Song, ``Contrastive adversarial learning for person
  independent facial emotion recognition,'' in \emph{Proceedings of the AAAI
  Conference on Artificial Intelligence}, vol.~35, no.~7, 2021, pp. 5948--5956.

\bibitem{kuruvayil2022emotion}
S.~Kuruvayil and S.~Palaniswamy, ``Emotion recognition from facial images with
  simultaneous occlusion, pose and illumination variations using
  meta-learning,'' \emph{Journal of King Saud University-Computer and
  Information Sciences}, vol.~34, no.~9, pp. 7271--7282, 2022.

\bibitem{rana2016emotion}
R.~Rana, R.~Jurdak, X.~Li, J.~Soar, R.~Goecke, J.~Epps, and M.~Breakspear,
  ``Emotion classification from noisy speech-a deep learning approach,''
  \emph{arXiv preprint arXiv:1603.05901}, 2016.

\bibitem{chatziagapi2019data}
A.~Chatziagapi, G.~Paraskevopoulos, D.~Sgouropoulos, G.~Pantazopoulos,
  M.~Nikandrou, T.~Giannakopoulos, A.~Katsamanis, A.~Potamianos, and
  S.~Narayanan, ``Data augmentation using gans for speech emotion
  recognition.'' in \emph{Interspeech}, 2019, pp. 171--175.

\bibitem{liu2020cross}
J.~Liu, W.~Zheng, Y.~Zong, C.~Lu, and C.~Tang, ``Cross-corpus speech emotion
  recognition based on deep domain-adaptive convolutional neural network,''
  \emph{IEICE TRANSACTIONS on Information and Systems}, vol. 103, no.~2, pp.
  459--463, 2020.

\bibitem{zadeh2017tensor}
A.~Zadeh, M.~Chen, S.~Poria, E.~Cambria, and L.-P. Morency, ``Tensor fusion
  network for multimodal sentiment analysis,'' \emph{arXiv preprint
  arXiv:1707.07250}, 2017.

\bibitem{zhou2020graph}
J.~Zhou, G.~Cui, S.~Hu, Z.~Zhang, C.~Yang, Z.~Liu, L.~Wang, C.~Li, and M.~Sun,
  ``Graph neural networks: A review of methods and applications,'' \emph{AI
  open}, vol.~1, pp. 57--81, 2020.

\bibitem{busso2008iemocap}
C.~Busso, M.~Bulut, C.-C. Lee, A.~Kazemzadeh, E.~Mower, S.~Kim, J.~N. Chang,
  S.~Lee, and S.~S. Narayanan, ``Iemocap: Interactive emotional dyadic motion
  capture database,'' \emph{Language resources and evaluation}, vol.~42, pp.
  335--359, 2008.

\bibitem{poria2018meld}
S.~Poria, D.~Hazarika, N.~Majumder, G.~Naik, E.~Cambria, and R.~Mihalcea,
  ``Meld: A multimodal multi-party dataset for emotion recognition in
  conversations,'' \emph{arXiv preprint arXiv:1810.02508}, 2018.

\bibitem{zadeh2016mosi}
A.~Zadeh, R.~Zellers, E.~Pincus, and L.-P. Morency, ``Mosi: multimodal corpus
  of sentiment intensity and subjectivity analysis in online opinion videos,''
  \emph{arXiv preprint arXiv:1606.06259}, 2016.

\bibitem{siddhant2019unsupervised}
A.~Siddhant, A.~Goyal, and A.~Metallinou, ``Unsupervised transfer learning for
  spoken language understanding in intelligent agents,'' in \emph{Proceedings
  of the AAAI conference on artificial intelligence}, vol.~33, no.~01, 2019,
  pp. 4959--4966.

\bibitem{medina2020self}
C.~Medina, A.~Devos, and M.~Grossglauser, ``Self-supervised prototypical
  transfer learning for few-shot classification,'' \emph{arXiv preprint
  arXiv:2006.11325}, 2020.

\bibitem{mao2020survey}
H.~H. Mao, ``A survey on self-supervised pre-training for sequential transfer
  learning in neural networks,'' \emph{arXiv preprint arXiv:2007.00800}, 2020.

\bibitem{ueno2019multi}
S.~Ueno, M.~Mimura, S.~Sakai, and T.~Kawahara, ``Multi-speaker
  sequence-to-sequence speech synthesis for data augmentation in
  acoustic-to-word speech recognition,'' in \emph{ICASSP 2019-2019 IEEE
  International Conference on Acoustics, Speech and Signal Processing
  (ICASSP)}.\hskip 1em plus 0.5em minus 0.4em\relax IEEE, 2019, pp. 6161--6165.

\bibitem{malik2023preliminary}
I.~Malik, S.~Latif, R.~Jurdak, and B.~Schuller, ``A preliminary study on
  augmenting speech emotion recognition using a diffusion model,'' \emph{arXiv
  preprint arXiv:2305.11413}, 2023.

\bibitem{latif2023generative}
S.~Latif, A.~Shahid, and J.~Qadir, ``Generative emotional ai for speech emotion
  recognition: The case for synthetic emotional speech augmentation,''
  \emph{Applied Acoustics}, vol. 210, p. 109425, 2023.

\bibitem{latif2018adversarial}
S.~Latif, R.~Rana, and J.~Qadir, ``Adversarial machine learning and speech
  emotion recognition: Utilizing generative adversarial networks for
  robustness,'' \emph{arXiv preprint arXiv:1811.11402}, 2018.

\bibitem{wolf2020transformers}
T.~Wolf, L.~Debut, V.~Sanh, J.~Chaumond, C.~Delangue, A.~Moi, P.~Cistac,
  T.~Rault, R.~Louf, M.~Funtowicz \emph{et~al.}, ``Transformers:
  State-of-the-art natural language processing,'' in \emph{Proceedings of the
  2020 conference on empirical methods in natural language processing: system
  demonstrations}, 2020, pp. 38--45.

\bibitem{vaswani2017attention}
A.~Vaswani, N.~Shazeer, N.~Parmar, J.~Uszkoreit, L.~Jones, A.~N. Gomez,
  {\L}.~Kaiser, and I.~Polosukhin, ``Attention is all you need,'' in
  \emph{Proceedings of the 31st International Conference on Neural Information
  Processing Systems}, 2017, pp. 6000--6010.

\bibitem{devlin2018bert}
J.~Devlin, M.-W. Chang, K.~Lee, and K.~Toutanova, ``Bert: Pre-training of deep
  bidirectional transformers for language understanding,'' \emph{arXiv preprint
  arXiv:1810.04805}, 2018.

\bibitem{radford2019language}
A.~Radford, J.~Wu, R.~Child, D.~Luan, D.~Amodei, and I.~Sutskever, ``Language
  models are unsupervised multitask learners,'' OpenAI, Tech. Rep., 2019.

\bibitem{chen2022key}
W.~Chen, X.~Xing, X.~Xu, J.~Yang, and J.~Pang, ``Key-sparse transformer for
  multimodal speech emotion recognition,'' in \emph{ICASSP 2022-2022 IEEE
  International Conference on Acoustics, Speech and Signal Processing
  (ICASSP)}.\hskip 1em plus 0.5em minus 0.4em\relax IEEE, 2022, pp. 6897--6901.

\bibitem{abdullah2021paralinguistic}
R.~M. Abdullah, S.~Y. Ameen, D.~M. Ahmed, S.~F. Kak, H.~M. Yasin, I.~M.
  Ibrahim, A.~M. Ahmed, Z.~N. Rashid, N.~Omar, and A.~A. Salih,
  ``Paralinguistic speech processing: An overview,'' \emph{Asian Journal of
  Research in Computer Science}, pp. 34--46, 2021.

\bibitem{wagner2023dawn}
J.~Wagner, A.~Triantafyllopoulos, H.~Wierstorf, M.~Schmitt, F.~Burkhardt,
  F.~Eyben, and B.~W. Schuller, ``Dawn of the transformer era in speech emotion
  recognition: closing the valence gap,'' \emph{IEEE Transactions on Pattern
  Analysis and Machine Intelligence}, 2023.

\bibitem{zenkov2021transformer}
I.~Zenkov, ``Transformer-cnn emotion recognition,''
  \url{https://github.com/IliaZenkov/transformer-cnn-emotion-recognition},
  2021.

\bibitem{li2021speech}
Y.~Li, Z.~Li, Z.~Zhang, X.~Li, and J.~Li, ``Speech emotion recognition
  transformer: A novel end-to-end model for ser,'' \emph{Neurocomputing}, vol.
  454, pp. 1--10, 2021.

\bibitem{park2019msp}
J.~Park and C.~Busso, ``Msp-podcast: A large-scale dataset of natural and
  emotionally evocative speech,'' in \emph{2019 IEEE Automatic Speech
  Recognition and Understanding Workshop (ASRU)}.\hskip 1em plus 0.5em minus
  0.4em\relax IEEE, 2019, pp. 112--119.

\bibitem{triantafyllopoulos2022probing}
A.~Triantafyllopoulos, J.~Wagner, H.~Wierstorf, M.~Schmitt, U.~Reichel,
  F.~Eyben, F.~Burkhardt, and B.~W. Schuller, ``Probing speech emotion
  recognition transformers for linguistic knowledge,'' in \emph{Proc.
  Interspeech 2022}, 2022, pp. 146--150.

\bibitem{chen2019complementary}
F.~Chen, Z.~Luo, Y.~Xu, and D.~Ke, ``Complementary fusion of multi-features and
  multi-modalities in sentiment analysis,'' \emph{arXiv preprint
  arXiv:1904.08138}, 2019.

\bibitem{huang2020multimodal}
J.~Huang, J.~Tao, B.~Liu, Z.~Lian, and M.~Niu, ``Multimodal transformer fusion
  for continuous emotion recognition,'' in \emph{ICASSP 2020-2020 IEEE
  International Conference on Acoustics, Speech and Signal Processing
  (ICASSP)}.\hskip 1em plus 0.5em minus 0.4em\relax IEEE, 2020, pp. 3507--3511.

\bibitem{siriwardhana2020multimodal}
S.~Siriwardhana, T.~Kaluarachchi, M.~Billinghurst, and S.~Nanayakkara,
  ``Multimodal emotion recognition with transformer-based self supervised
  feature fusion,'' \emph{IEEE Access}, vol.~8, pp. 176\,274--176\,285, 2020.

\bibitem{su2020improving}
Z.~Su, Z.~Zhang, X.~Li, and J.~Li, ``Improving speech emotion recognition using
  graph attentive bi-directional gated recurrent unit network,'' in \emph{Proc.
  Interspeech 2020}, 2020, pp. 4861--4865.

\bibitem{wang2021learning}
Y.~Wang, G.~Shen, Y.~Xu, J.~Li, and Z.~Zhao, ``Learning mutual correlation in
  multimodal transformer for speech emotion recognition.'' in
  \emph{Interspeech}, 2021, pp. 4518--4522.

\bibitem{zheng2021fused}
R.~Zheng, J.~Chen, M.~Ma, and L.~Huang, ``Fused acoustic and text encoding for
  multimodal bilingual pretraining and speech translation,'' in
  \emph{International Conference on Machine Learning}.\hskip 1em plus 0.5em
  minus 0.4em\relax PMLR, 2021, pp. 12\,736--12\,746.

\bibitem{kim2022representation}
J.~Kim and J.~Kim, ``Representation learning with graph neural networks for
  speech emotion recognition,'' \emph{arXiv preprint arXiv:2208.09830}, 2022.

\bibitem{zhang2022transformer}
W.~Zhang, F.~Qiu, S.~Wang, H.~Zeng, Z.~Zhang, R.~An, B.~Ma, and Y.~Ding,
  ``Transformer-based multimodal information fusion for facial expression
  analysis,'' in \emph{Proceedings of the IEEE/CVF Conference on Computer
  Vision and Pattern Recognition}, 2022, pp. 2428--2437.

\bibitem{guo2022emotion}
L.~Guo, L.~Wang, J.~Dang, Y.~Fu, J.~Liu, and S.~Ding, ``Emotion recognition
  with multimodal transformer fusion framework based on acoustic and lexical
  information,'' \emph{IEEE MultiMedia}, vol.~29, no.~2, pp. 94--103, 2022.

\bibitem{yang2022self}
K.~Yang, R.~Zhang, H.~Xu, and K.~Gao, ``A self-adjusting fusion representation
  learning model for unaligned text-audio sequences,'' \emph{arXiv preprint
  arXiv:2212.11772}, 2022.

\bibitem{wang2023multimodal}
Y.~Wang, Y.~Gu, Y.~Yin, Y.~Han, H.~Zhang, S.~Wang, C.~Li, and D.~Quan,
  ``Multimodal transformer augmented fusion for speech emotion recognition,''
  \emph{Frontiers in Neurorobotics}, vol.~17, p. 1181598, 2023.

\bibitem{dutta2023hcam}
S.~Dutta and S.~Ganapathy, ``Hcam--hierarchical cross attention model for
  multi-modal emotion recognition,'' \emph{arXiv preprint arXiv:2304.06910},
  2023.

\bibitem{xie2021robust}
B.~Xie, M.~Sidulova, and C.~H. Park, ``Robust multimodal emotion recognition
  from conversation with transformer-based crossmodality fusion,''
  \emph{Sensors}, vol.~21, no.~14, p. 4913, 2021.

\bibitem{schlegel2021training}
K.~Schlegel, K.~R. Scherer, and M.~Mortillaro, ``Training emotion recognition
  accuracy: Results for multimodal expressions and micro expressions,''
  \emph{Frontiers in Psychology}, vol.~12, p. 708867, 2021.

\bibitem{wang2022multi}
Q.~Wang, M.~Wang, Y.~Yang, and X.~Zhang, ``Multi-modal emotion recognition
  using eeg and speech signals,'' \emph{Computers in Biology and Medicine},
  vol. 149, p. 105907, 2022.

\bibitem{tsai2019multimodal}
Y.-H.~H. Tsai, S.~Bai, P.~P. Liang, J.~Z. Kolter, L.-P. Morency, and
  R.~Salakhutdinov, ``Multimodal transformer for unaligned multimodal language
  sequences,'' in \emph{Proceedings of the conference. Association for
  Computational Linguistics. Meeting}, vol. 2019.\hskip 1em plus 0.5em minus
  0.4em\relax NIH Public Access, 2019, p. 6558.

\bibitem{liu2021multi}
D.~Liu, Z.~Wang, L.~Wang, and L.~Chen, ``Multi-modal fusion emotion recognition
  method of speech expression based on deep learning,'' \emph{Frontiers in
  Neurorobotics}, vol.~15, p. 697634, 2021.

\bibitem{liu2022multi}
Y.~Liu, H.~Sun, W.~Guan, Y.~Xia, and Z.~Zhao, ``Multi-modal speech emotion
  recognition using self-attention mechanism and multi-scale fusion
  framework,'' \emph{Speech Communication}, vol. 139, pp. 1--9, 2022.

\bibitem{ho2020multi}
C.-P. Ho, C.-C. Yang, S.~Kim, and Y.-N. Lee, ``Multi-head attention fusion
  networks for multi-modal speech emotion recognition,'' \emph{Computer Speech
  \& Language}, vol.~65, p. 101122, 2020.

\bibitem{morais2022speech}
E.~Morais, R.~Hoory, W.~Zhu, I.~Gat, M.~Damasceno, and H.~Aronowitz, ``Speech
  emotion recognition using self-supervised features,'' in \emph{ICASSP
  2022-2022 IEEE International Conference on Acoustics, Speech and Signal
  Processing (ICASSP)}.\hskip 1em plus 0.5em minus 0.4em\relax IEEE, 2022, pp.
  6922--6926.

\bibitem{tang2022multimodal}
W.~Tang, F.~He, Y.~Liu, and Y.~Duan, ``Matr: Multimodal medical image fusion
  via multiscale adaptive transformer,'' \emph{IEEE Transactions on Image
  Processing}, vol.~31, pp. 5134--5149, 2022.

\bibitem{yoon2018multimodal}
D.~Yoon, S.~Lee, and H.~Lee, ``Multimodal speech emotion recognition using
  audio and text,'' in \emph{2018 IEEE Spoken Language Technology Workshop
  (SLT)}.\hskip 1em plus 0.5em minus 0.4em\relax IEEE, 2018, pp. 112--118.

\bibitem{jin2021hybrid}
Z.~Jin, J.~Li, and J.~Tang, ``Hybrid transformer: a flexible and efficient
  neural network for multimodal applications,'' in \emph{Proceedings of the
  2021 Conference of the North American Chapter of the Association for
  Computational Linguistics: Human Language Technologies}, 2021, pp.
  2870--2882.

\bibitem{burkhardt2005database}
F.~Burkhardt, A.~Paeschke, M.~Rolfes, W.~F. Sendlmeier, and B.~Weiss, ``A
  database of german emotional speech.'' in \emph{Interspeech}, vol.~5, 2005,
  pp. 1517--1520.

\bibitem{costantini2014emovo}
G.~Costantini, I.~Iaderola, A.~Paoloni, M.~Todisco \emph{et~al.}, ``Emovo
  corpus: an italian emotional speech database,'' in \emph{Proceedings of the
  ninth international conference on language resources and evaluation
  (LREC'14)}.\hskip 1em plus 0.5em minus 0.4em\relax European Language
  Resources Association (ELRA), 2014, pp. 3501--3504.

\bibitem{latif2022self}
S.~Latif, R.~Rana, S.~Khalifa, R.~Jurdak, and B.~W. Schuller, ``Self supervised
  adversarial domain adaptation for cross-corpus and cross-language speech
  emotion recognition,'' \emph{IEEE Transactions on Affective Computing}, 2022.

\bibitem{xu2019learning}
H.~Xu, H.~Zhang, K.~Han, Y.~Wang, Y.~Peng, and X.~Li, ``Learning alignment for
  multimodal emotion recognition from speech,'' \emph{arXiv preprint
  arXiv:1909.05645}, 2019.

\bibitem{sebastian2019fusion}
J.~Sebastian, P.~Pierucci \emph{et~al.}, ``Fusion techniques for
  utterance-level emotion recognition combining speech and transcripts.'' in
  \emph{Interspeech}, 2019, pp. 51--55.

\bibitem{chen2020multi}
M.~Chen and X.~Zhao, ``A multi-scale fusion framework for bimodal speech
  emotion recognition.'' in \emph{Interspeech}, 2020, pp. 374--378.

\bibitem{krishna2020multimodal}
D.~Krishna and A.~Patil, ``Multimodal emotion recognition using cross-modal
  attention and 1d convolutional neural networks.'' in \emph{Interspeech},
  2020, pp. 4243--4247.

\bibitem{sun2021multimodal}
L.~Sun, B.~Liu, J.~Tao, and Z.~Lian, ``Multimodal cross-and self-attention
  network for speech emotion recognition,'' in \emph{ICASSP 2021-2021 IEEE
  International Conference on Acoustics, Speech and Signal Processing
  (ICASSP)}.\hskip 1em plus 0.5em minus 0.4em\relax IEEE, 2021, pp. 4275--4279.

\bibitem{lian2021ctnet}
Z.~Lian, B.~Liu, and J.~Tao, ``Ctnet: Conversational transformer network for
  emotion recognition,'' \emph{IEEE/ACM Transactions on Audio, Speech, and
  Language Processing}, vol.~29, pp. 985--1000, 2021.

\bibitem{kumar2021towards}
P.~Kumar, V.~Kaushik, and B.~Raman, ``Towards the explainability of multimodal
  speech emotion recognition.'' in \emph{Interspeech}, 2021, pp. 1748--1752.

\bibitem{latif2018urdu}
S.~Latif, R.~Rana, J.~Qadir, and J.~Epps, ``Urdu speech corpus for emotion
  recognition,'' \emph{Data in brief}, vol.~19, pp. 2187--2191, 2018.

\bibitem{schuller2010cross}
B.~Schuller, S.~Steidl, A.~Batliner, F.~Burkhardt, L.~Devillers, C.~M{\"u}ller,
  and S.~Narayanan, ``Cross-corpus acoustic emotion recognition: Variances and
  strategies,'' in \emph{2010 IEEE International Conference on Acoustics,
  Speech and Signal Processing}.\hskip 1em plus 0.5em minus 0.4em\relax IEEE,
  2010, pp. 5118--5121.

\bibitem{baevski2020wav2vec}
A.~Baevski, Y.~Zhou, A.~Mohamed, and M.~Auli, ``wav2vec 2.0: A framework for
  self-supervised learning of speech representations,'' \emph{Advances in
  Neural Information Processing Systems}, vol.~33, pp. 12\,449--12\,460, 2020.

\bibitem{2108.09669}
K.~D. N, ``Using large pre-trained models with cross-modal attention for
  multi-modal emotion recognition,'' 2021.

\bibitem{hochreiter1997long}
S.~Hochreiter and J.~Schmidhuber, ``Long short-term memory,'' \emph{Neural
  computation}, vol.~9, no.~8, 1997.

\end{thebibliography}
\end{document}